\documentclass{article}

\usepackage[preprint]{neurips_2026}
\usepackage{amsmath}

\usepackage[utf8]{inputenc} 
\usepackage[T1]{fontenc}    
\usepackage{hyperref}       
\usepackage{url}            
\usepackage{booktabs}       
\usepackage{amsfonts}       
\usepackage{nicefrac}       
\usepackage{microtype}      
\usepackage{xcolor}         

\usepackage{comment}
\usepackage[caption=false]{subfig}
\usepackage{graphicx}
\usepackage{textcomp}
\usepackage{tabularx}
\usepackage{geometry}
\usepackage{multirow}
\usepackage{caption}
\usepackage{lipsum}
\usepackage{colortbl} 
\usepackage{enumitem}
\usepackage{pifont}
\usepackage{subcaption}
\usepackage{setspace}

\usepackage{fontawesome5}

\title{EEG-X: Toward Device-Agnostic and Noise-Robust Foundation Models for EEG}

\author{%
  Navid Mohammadi Foumani \\
  Emotiv Research \\
  Melbourne, Australia \\
  \texttt{navid@emotiv.com}
  \And
  Soheila Ghane\thanks{Work done while at Emotiv Research.} \\
  Google \\
  Melbourne, Australia \\
  \texttt{soheila@google.com}
  \And
  Nam Nguyen \\
  Emotiv Research \\
  Sydney, Australia \\
  \texttt{namnguyen@emotiv.com}
  \AND
  Mahsa Salehi \\
  Monash University \\
  Melbourne, Australia \\
  \texttt{mahsa.salehi@monash.edu}
  \And
  Geoffrey I.~Webb \\
  Monash University \\
  Melbourne, Australia \\
  \texttt{geoff.webb@monash.edu}
  \And
  Geoffrey Mackellar \\
  Emotiv Research \\
  Sydney, Australia \\
  \texttt{geoff@emotiv.com}
}

\begin{document}

\maketitle

\begin{abstract}
Foundation models for EEG analysis are still in their infancy, limited by two key challenges: (1) variability across datasets caused by differences in recording devices and electrode configurations, and (2) the low signal-to-noise ratio (SNR) of EEG, where neural activity is often buried under artifacts and non-brain sources. To address these challenges, we present \textit{EEG-X}, a device-agnostic and noise-robust foundation model for EEG representation learning. EEG-X introduces a location-based channel embedding that enables robust transfer across heterogeneous EEG devices and electrode layouts. To improve robustness against noise, EEG-X employs a novel noise-aware masking-reconstruction strategy that reconstructs artifact-cleaned signals rather than raw noisy EEG and introduces a novel Dictionary-based Convolution Transformation (DiCT) that improves reconstruction-based pretraining by comparing signals in a structured feature space instead of directly in the raw signal space. Experiments across datasets collected from diverse EEG devices show that EEG-X consistently outperforms state-of-the-art methods across multiple downstream tasks and demonstrates strong cross-domain generalization when pretraining and downstream datasets differ in electrode layouts and recording configurations paving the way towards foundation models. The models and code are available at ~\href{https://github.com/Emotiv/EEG-X}{\texttt{github.com/Emotiv/EEG-X}} \faGithub.
\end{abstract}

\section{Introduction}
Electroencephalography (EEG) is a non-invasive method for recording brain activity by placing electrodes on the scalp, offering high temporal resolution and relatively low cost~\citep{EEG2002,EEG2020}. It is widely used in both neuroscience and clinical practice, where it supports the diagnosis of neurological disorders such as epilepsy and sleep disorders~\citep{EEG18App, EEG20seizure}, as well as in applications like sleep stage classification, affective state analysis, motor imagery decoding, stress detection, and seizure classification~\citep{EEG19App,EEG20App,rakovic2024measuring}. Despite these benefits, effectively modeling EEG data often requires large amounts of annotated data, expert knowledge, and techniques specially designed to the unique characteristics of each task~\citep{weng2024self}.

Recently, self-supervised pretraining has achieved remarkable success in natural language processing (NLP), computer vision (CV), and speech/audio processing~\citep{bommasani2021opportunities,yu2022coca,chen2024vast}. Models pretrained on large unlabeled datasets can be efficiently adapted to a wide range of downstream tasks, often requiring little or no labeled data. A major advantage of these models is their ability to handle multiple tasks within a single framework, unlike traditional approaches that typically require large labeled datasets and task-specific algorithms. This representation learning approach simplifies development by reducing the need for task-specific models and extensive labeling, while also improving generalization by leveraging shared patterns from diverse data. This makes the model more adaptable and efficient across various tasks, forming the basis of what is known as a foundation model.

While foundation models have shown great success in vision, language, and speech, their application to EEG data remains limited and has yet to achieve comparable breakthroughs~\citep{weng2024self,foumani2023deep}. Two key challenges hinder the broader adoption of EEG foundation models: (1) substantial variability across EEG datasets due to differences in recording equipment, which introduce variations in sampling frequencies, electrode placements, and the number of channels, and (2) the inherently low signal-to-noise ratio of EEG, where neural activity is often buried under artifacts and non-brain sources~\citep{artifacts_survey,IClabel,rASR}. To address these challenges, we propose EEG-X, an EEG representation learning model that learns device-agnostic and noise-robust representations through a location-based channel embedding to handle variability across devices, a noise-aware masking–reconstruction strategy for robust pretraining, and a dictionary-based convolution transformation layer that improves reconstruction loss by enabling a balanced frequency- and shape-aware comparisons.

The first component, location-based channel embedding, is designed to handle device variability. Earlier models typically rely on learnable channel embeddings~\citep{Biot,LaBraM,EEGPT,cbramod,steegformer} to indicate signal source, which can work within a single dataset but transfer poorly across devices, as the same channels may not exist across different EEG devices. 
In addition, such embeddings do not account for electrode positions or preserve the similarity between neighboring regions on the scalp, both of which are important for modeling brain activity. EEG-X instead encodes electrode positions together with their local spatial neighborhoods, enabling nearby electrodes to share similar representations. Furthermore, EEG-X introduces a universal scalp mapping strategy based on a dense scalp mesh spanning the standard 10-05, 10-10, and 10-20 EEG systems, allowing electrodes from heterogeneous devices to be mapped into a shared spatial representation.

The second component addresses the low Signal-to-Noise Ratio (low-SNR) nature of EEG signals. Unlike previous reconstruction-based methods that reconstruct the original noisy EEG signals~\citep{Bendr,maeeg,EEGPT,cbramod,reve,steegformer}, EEG-X reconstructs artifact-cleaned signals obtained through an artifact removal process. This encourages the model to focus on neural activity rather than noise and improves the robustness and quality of the learned representations.

Finally, EEG-X introduces a \textbf{Di}ctionary-based \textbf{C}onvolution \textbf{T}ransformation (DiCT) layer to improve the reconstruction objective in MAE-based architectures. While Mean Squared Error (MSE) is widely used in self-supervised prediction tasks~\citep{data2vec2-2023,data2vec2022}, it may be suboptimal for EEG signals due to its point-wise nature and tendency to overemphasize large-amplitude components~\citep{MSE,foumani2024series2vec}. Consequently, MSE may inadequately capture frequency-specific and structural similarities in EEG signals. To address these limitations, DiCT projects both reconstructed and target EEG signals into a structured feature space using random convolutional kernels with multiple dilation scales before computing reconstruction loss. By computing MSE on projected temporal features rather than directly in the raw signal space, DiCT improves sensitivity to multi-scale temporal patterns while increasing robustness to amplitude variation and temporal misalignment. Importantly, DiCT has no learnable parameters and can be seamlessly integrated into existing reconstruction-based methods. Here is a summary of our contributions:

\begin{itemize}[left=0.6em]
\item \textbf{Location-Based Channel Embedding}: Introduces a spatially-aware channel embedding that encodes electrode locations and local scalp neighborhoods, enabling robust transfer across heterogeneous EEG devices and channel layouts.
\item \textbf{Noise-Aware Reconstruction}: Proposes a reconstruction strategy based on artifact-cleaned EEG signals, encouraging representations to focus on neural activity while improving robustness to low-SNR EEG recordings.
\item \textbf{Dictionary-based Convolution Transformation (DiCT)}: Introduces a lightweight structured reconstruction objective that mitigates limitations of point-wise MSE by improving sensitivity to multi-scale frequency content and temporal structure.
\item \textbf{Extensive Evaluation}: Demonstrates that EEG-X serves as a strong representation learning model for EEG, achieving state-of-the-art performance across diverse downstream tasks and datasets while exhibiting robust cross-domain generalization across heterogeneous EEG devices and recording configurations.
\end{itemize}

\section{Methodology of EEG-X}

\begin{figure}[]
    \centering
    \includegraphics[trim=0.5cm 0.5cm 0.5cm 0cm, width=\textwidth]{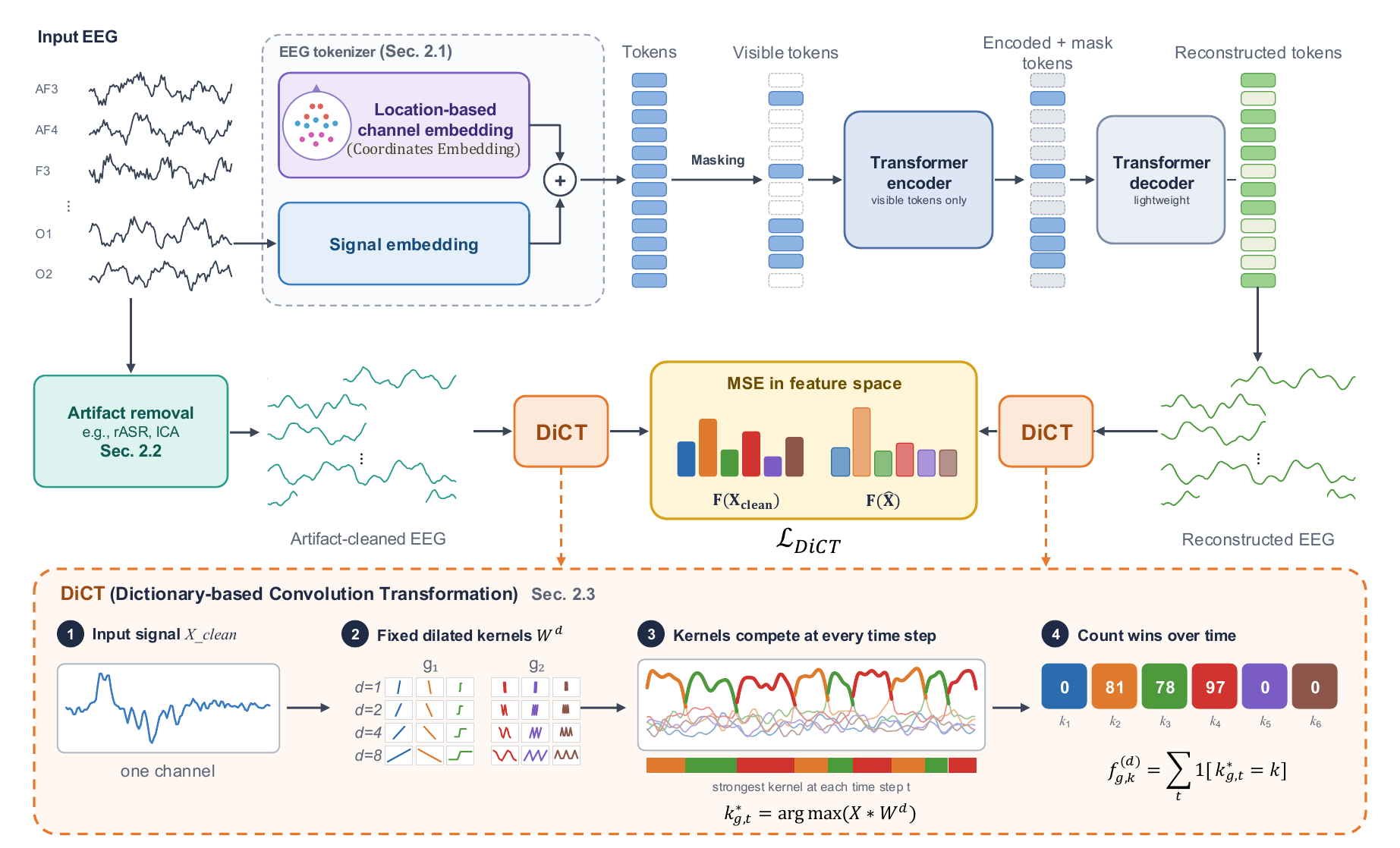}
    \caption{Architecture of EEG-X. EEG-X is built upon a masked autoencoding (MAE) framework and consists of three main components: (1) a signal embedding module augmented with location-based channel embeddings for handling heterogeneous EEG devices and channel layouts, (2) an artifact removal module that produces artifact-cleaned reconstruction targets from raw EEG signals, and (3) Dictionary-based Convolution Transformation (DiCT) objective for structured reconstruction learning.}
    \label{fig:EEG-X}
\end{figure}  


Fig.~\ref{fig:EEG-X} illustrates the overall architecture of EEG-X, which consists of three main components: (1) an input tokenizer with location-based channel embedding, (2) a noise-aware masked reconstruction framework, and (3) a Dictionary-based Convolution Transformation (DiCT) reconstruction objective. Given an EEG signal, EEG-X first divides the input into fixed-length temporal segments and projects them into token representations using a signal embedding module. To incorporate spatial information, each channel is augmented with a location-based channel embedding derived from electrode coordinates and local scalp neighborhoods, enabling robust representation learning across heterogeneous EEG devices and channel layouts.

The resulting tokens are masked and processed using a Masked Autoencoder (MAE)~\citep{mae2022} framework, where the model reconstructs masked EEG tokens using contextual information from the remaining visible tokens. Similar to vision MAE frameworks, the encoder operates only on visible tokens to efficiently learn contextual representations, while the decoder reconstructs the missing regions from latent representations together with mask tokens. EEG-X adopts an asymmetric encoder--decoder design, where the decoder is intentionally smaller than the encoder to reduce computational cost and encourage the encoder to learn richer representations. 

Unlike previous reconstruction-based methods that reconstruct the original noisy EEG, EEG-X reconstructs artifact-cleaned EEG obtained through an artifact removal process, encouraging the model to focus on neural activity rather than noise. To further improve reconstruction learning, EEG-X introduces a Dictionary-based Convolution Transformation (DiCT) objective. Instead of computing reconstruction loss directly in the raw signal space, both the reconstructed and target EEG signals are projected into a structured feature space using random convolutional kernels with multiple dilation scales before computing MSE loss. By comparing the occurrence and distribution of temporal patterns rather than enforcing strict point-wise alignment, DiCT improves sensitivity to multi-scale temporal patterns while increasing robustness to amplitude variation and temporal misalignment. For downstream tasks, the tokenizer and encoder are frozen, and only a lightweight classifier is trained on top of the learned EEG representations.

\subsection{EEG Tokenizer} \label{sec:tokenizer}
Our goal is to model heterogeneous EEG signals—recorded using different devices with varying channel configurations, combinations, and recording lengths—within a unified framework.

\textbf{Signal Embedding:} 
Let $\mathbf{X}=\{x_1,\dots,x_C\}$ denote an EEG recording with $C$ channels, where each channel $x_c\in\mathbb{R}^{L}$ has $L$ samples. To handle recordings of different lengths, we split each channel into fixed-length tokens of $w$ samples using a sliding window with stride $s=w-o$, where $o<w$ is the overlap (typically $o=w/4$). The $i$-th token of channel $c$ is
\begin{equation}
t_{c,i}=x_c\big[(i-1)s+1 : (i-1)s+w\big],\qquad i=1,\dots,N,
\end{equation}
where $N=\lfloor (L-w)/s\rfloor+1$ is the number of tokens per channel.
To capture both temporal and spectral characteristics of EEG signals, we combine raw time-domain and frequency-domain representations for each token. Specifically, we apply the Short-Time Fourier Transform (STFT) to each token and extract the magnitude of its frequency representation~\citep{Biot}. The raw token and its corresponding STFT magnitude are then independently projected into embedding spaces using learned linear transformations. The resulting representations are concatenated to obtain the final token embedding $e_{c,i}$:
\begin{equation}
e_{c,i} =
\left[
W_{\mathrm{raw}} t_{c,i}
\ ; \
W_{\mathrm{freq}} \left|\mathrm{STFT}(t_{c,i})\right|
\right],
\quad e_{c,i} \in \mathbb{R}^{d_e}.
\end{equation}
Here, $\mathbb{R}^{d_e}$ denotes the embedding dimension, and the transformations are parameterized by learned weight matrices $W_{\mathrm{raw}}$ and $W_{\mathrm{freq}}$. The concatenation combines information from both the original temporal signal and its spectral representation, allowing the model to learn complementary features from the two domains. This tokenization process is similar to natural language processing, where each EEG sample is treated as a sequence of tokens, similar to words in a sentence.

While this signal embedding addresses variations in recording length, handling channel variability remains a challenge. Previous approaches train an embedding parameter for all channels in the data and add the corresponding channel embedding to the token~\citep{Biot,LaBraM,EEGPT,cbramod}; however, they present several limitations: (1) The same channels may not exist across different EEG devices, making the learned embeddings less transferable across diverse recording setups. (2) Channel embeddings may not account for the spatial locations of electrodes on the scalp, which are crucial for accurately modeling the brain's spatial activity patterns. (3) They do not preserve the similarity between neighboring regions on the scalp, which can lead to the loss of meaningful spatial relationships. These limitations highlight the need for a new embedding strategy that incorporates both the spatial location of the channels and their relationships across different devices, which our method aims to address.

\begin{figure*}
  \centering
  \subfloat[]{%
    \includegraphics[trim= 1cm 9cm 12cm 3.5cm, width=0.65\linewidth]{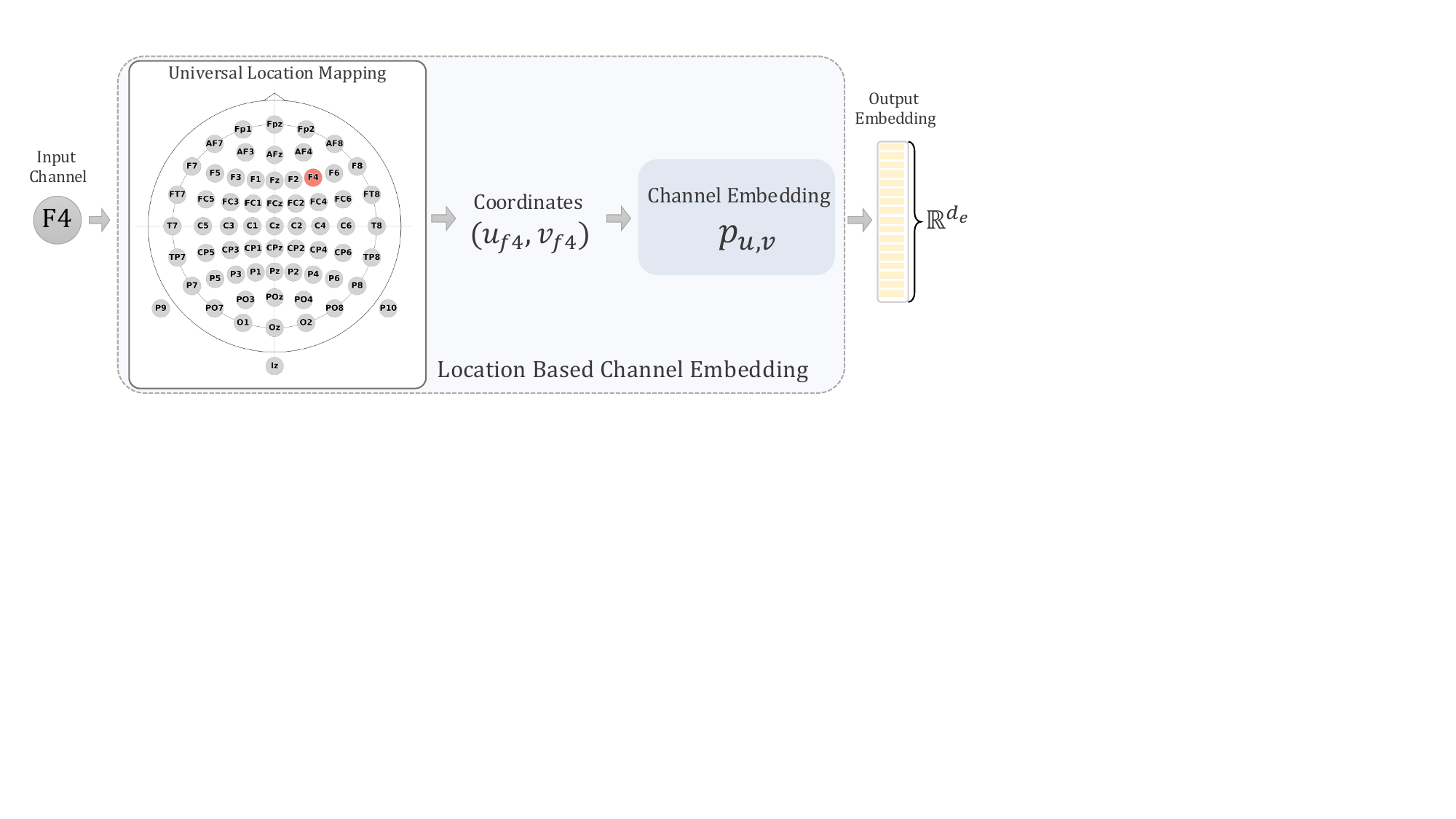}
    \label{fig:LCE}%
  }\hfill%
    \subfloat[]{%
    \includegraphics[trim=1cm 0cm 1cm 5.5cm, width=0.28\linewidth]{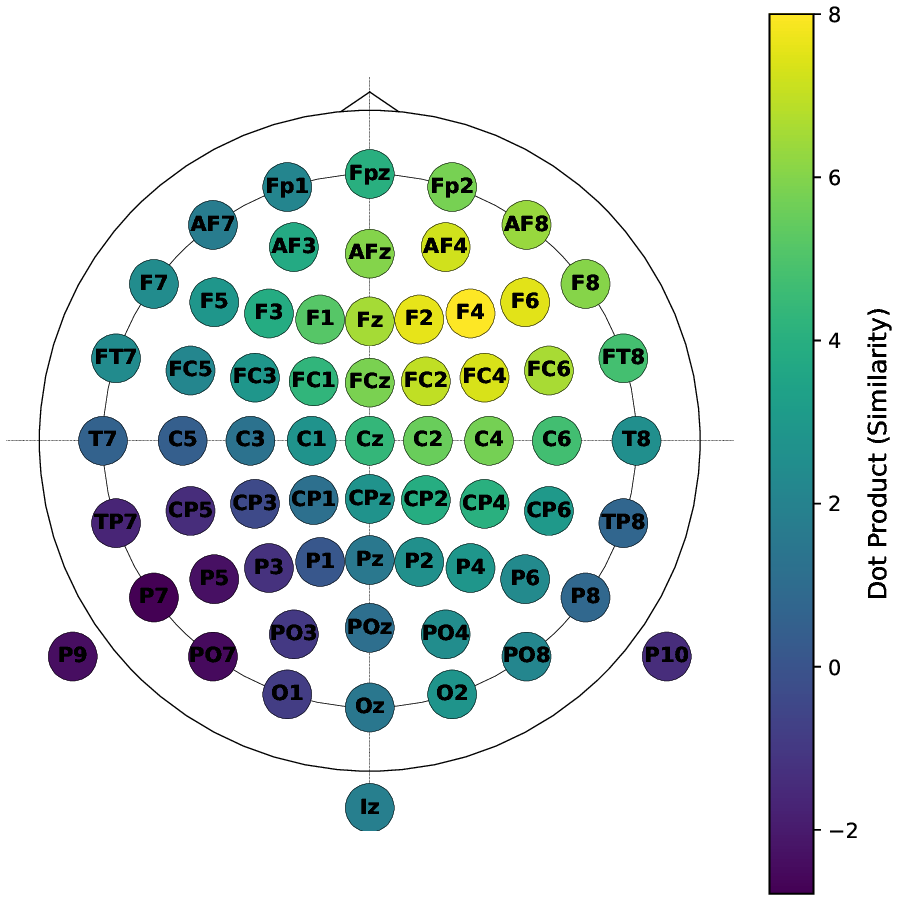}
    \label{fig:F4_position}%
  }

  \caption{(a) Location-based channel embedding using a universal scalp coordinate mapping to assign each channel Cartesian coordinates. (b) Embedding similarity of electrode F4 with all other channels, showing higher similarity to nearby electrodes (e.g., F2, F6) than distant ones (e.g., P7).}
  \label{fig:LCE_F4}

\end{figure*}

To address these limitations, EEG-X incorporates electrode location information directly into the token representations. Each electrode is mapped to Cartesian scalp coordinates using a universal scalp mapping (Fig.~\ref{fig:LCE}). For visualization purposes, Fig.~\ref{fig:LCE} illustrates a 64-channel configuration. In practice, EEG-X uses a substantially denser universal scalp mesh spanning standard 10-05, 10-10, and 10-20 EEG systems, enabling heterogeneous EEG devices to be represented within a shared spatial coordinate space (Appendix~\ref{apx:Location_mapping}). Recent work such as REVE~\citep{reve} also combines spatial location encoding with learnable channel embeddings. However, REVE employs a joint multidimensional Fourier encoding, whereas EEG-X adopts a simpler separable sinusoidal formulation together with the proposed universal scalp mapping. In practice, we find that the separable formulation is more robust and remains stable even at relatively small embedding dimensions.

Given an electrode location $(u,v)$, where $u$ and $v$ denote the horizontal and vertical coordinates (i.e., $x$ and $y$), we generate spatial embeddings using a separable sinusoidal encoding adapted from transformer positional embeddings~\citep{attention2017,ConvTran2023}:

\begin{equation}
\begin{aligned}
    p_{u,v}(4k) &= \mathrm{sin}\, u\omega_k, \quad 
    p_{u,v}(4k+1) = \mathrm{cos}\, u\omega_k, \\
    p_{u,v}(4k+2) &= \mathrm{sin}\, v\omega_k, \quad 
    p_{u,v}(4k+3) = \mathrm{cos}\, v\omega_k
\end{aligned}
\label{eq:apeV}
\end{equation}
where
\[
\omega_k = 1000^{-4k/d_e},
\]
$k \in \{0, \dots, d_e/4-1\}$, and $d_e$ is the embedding dimension. The frequency term $\omega_k$ assigns multiple spatial frequencies across embedding dimensions, enabling the model to represent both coarse and fine-grained spatial relationships between electrodes.

This approach allows signals from nearby electrode positions to receive similar embeddings. Specifically:
\begin{align}
\text{Dist}((u_i, v_i), (u_j, v_j)) 
&< \text{Dist}((u_i, v_i), (u_k, v_k)) \nonumber 
&\implies p_{u_i, v_i} \cdot p_{u_j, v_j} 
> p_{u_i, v_i} \cdot p_{u_k, v_k}
\end{align}
Where $\text{Dist()}$ is the distance function in Cartesian coordinates, and $p_{u_i, v_i}, p_{u_j, v_j} \in \mathbb{R}^{d_e}$ represent the channel embeddings of the electrodes at positions $(u_i, v_i)$ and $(u_j, v_j)$, respectively. The equation demonstrates that electrodes closer in physical space (i.e., neighboring electrodes) will have higher similarity in their embeddings, as indicated by the dot product.

Fig. \ref{fig:F4_position} shows the embedding similarity between electrods $F4$ with all other electrodes. For example, as shown in Fig. \ref{fig:F4_position}, signals from the $F4$ electrode will have embeddings more similar to those of $F2$ and $F6$ than to other electrodes (such as $P7$). As a result, when any of these electrodes are present in a dataset, the model can effectively utilize them. If an electrode is missing or too noisy to be useful, the model can reconstruct the signal primarily using data from adjacent electrodes. This approach mitigates the issue of missing channels across devices, as it maintains consistent channel relationships based on location, even if the channel labeling differs. Additionally, this encoding provides the model with spatial awareness, enabling it to recognize regional brain activity patterns, where neighboring areas exhibit similar behaviors while differing from other regions.

\subsection{Noise-Aware Reconstruction}
\label{sec:noise_reconstruction}

EEG signals exhibit inherently low signal-to-noise ratio (SNR), where neural activity is often contaminated by artifacts and non-brain sources. Previous MAE-based self-supervised methods primarily reconstruct the original raw EEG signal~\citep{maeeg,Biot,EEGPT,cbramod,reve,steegformer}; however, this is particularly challenging for EEG, as artifacts can dominate the underlying neural activity. Consequently, reconstruction objectives may encourage models to preserve artifact-related patterns together with meaningful brain signals, reducing the quality and robustness of the learned representations.

To address this limitation, EEG-X uses artifact-cleaned EEG signals as reconstruction targets instead of the original noisy input. Our framework is agnostic to the choice of denoising method; in this work, we adopt Riemannian Artifact Subspace Reconstruction (rASR)~\citep{rASR} due to its robustness, computational efficiency, and suitability for large-scale EEG processing (see Appendix~\ref{apx:rASR} for a broader discussion of alternative denoising methods). Given an input EEG signal $\mathbf{X}$, rASR produces an artifact-cleaned signal $\mathbf{X}_{\text{clean}}$, which serves as the reconstruction target during self-supervised pretraining. Detailed descriptions of the rASR calibration process, parameter selection, and implementation details are provided in Appendix~\ref{apx:rASR}. By reconstructing artifact-cleaned EEG rather than noisy raw signals, EEG-X encourages the learned representations to focus on neural activity while improving robustness to artifacts and recording variability.

\begin{figure}[h]
    \centering
    \includegraphics[trim=0.5cm 0.5cm 0cm 0cm, width=1\textwidth]{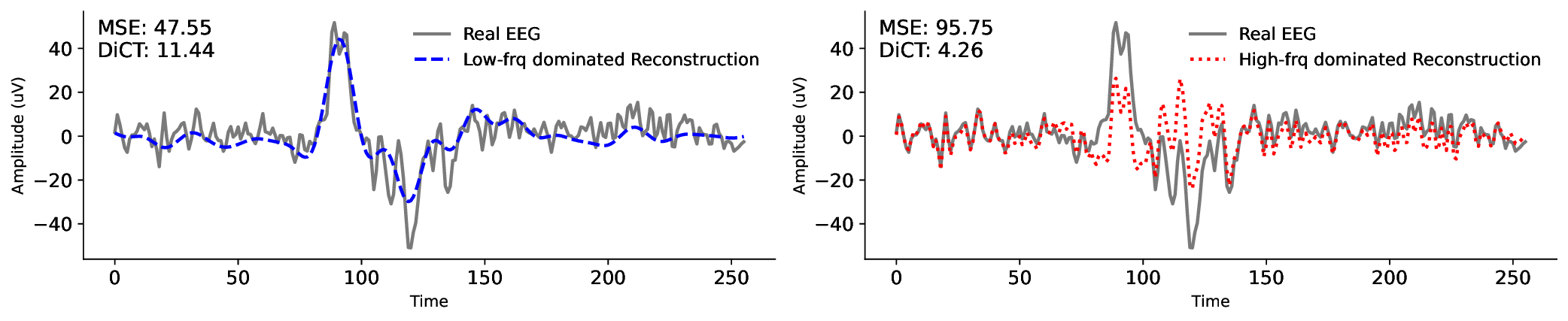}
    \caption{
    Real EEG signal and reconstructed variants. 
    \textbf{Left:} low-frequency dominant reconstruction preserving mainly Delta and Theta components. 
    \textbf{Right:} mid/high-frequency dominant reconstruction preserving Theta, Alpha, Beta, and Gamma components. 
    These examples illustrate the amplitude bias of direct time-domain MSE, where low-frequency high-amplitude components can dominate the reconstruction objective despite missing informative higher-frequency temporal patterns.
    }
    \label{fig:DiCT}

\end{figure}

\subsection{Dictionary-based Convolution Transformation (DiCT)} \label{DiCT}
As illustrated in Fig.~\ref{fig:EEG-X}, EEG-X is built upon a masked autoencoding (MAE) framework~\citep{mae2022}, where portions of the input EEG tokens are masked and the model is trained to reconstruct the missing regions from the remaining visible tokens. A straightforward approach is to compute reconstruction loss directly in the raw signal space using Mean Squared Error (MSE), as commonly adopted in previous self-supervised EEG methods~\citep{maeeg,Biot,EEGPT,cbramod,reve,steegformer}. However, EEG signals are inherently multi-scale, where meaningful neural activity is distributed across multiple temporal resolutions and frequency bands with substantially different amplitudes and dynamics.

EEG recordings are also highly susceptible to artifacts and non-brain sources. For example, eye blinks and movement artifacts often dominate low-frequency regions, while muscle activity and line noise introduce strong high-frequency components. As a result, both low- and high-frequency regions can be dominated by non-neural, high-amplitude noise, causing point-wise losses such as MSE to be disproportionately influenced by large-amplitude components. Fig.~\ref{fig:DiCT} illustrates this limitation using a real EEG signal together with two reconstructed variants. The left reconstruction preserves only low-frequency components, specifically Delta (1--4 Hz) and Theta (4--8 Hz) bands, while the right reconstruction preserves richer temporal structure including Theta (4--8 Hz), Alpha (8--14 Hz), Beta (14--30 Hz), and Gamma ($>$30 Hz) components (details of signal construction are provided in Appendix~\ref{apx:DiCT}). Despite the right reconstruction containing more informative temporal dynamics, direct time-domain MSE favors the low-frequency dominant reconstruction because its large-amplitude components align more closely with the original signal. Specifically, the low-frequency dominant reconstruction achieves a lower MSE (47.55) compared to the richer mid/high-frequency reconstruction (95.75).

To address these limitations, EEG-X introduces the \textbf{Di}ctionary-based \textbf{C}onvolution \textbf{T}ransformation (DiCT), which projects both the reconstructed and the target EEG signals into a structured feature space before computing the reconstruction loss. DiCT is inspired by the success of random convolutional kernels~\citep{Rocket} and competing kernel groups~\citep{Hydra} in time series classification. Instead of enforcing exact point-wise alignment, it compares the occurrence and distribution of temporal patterns across multiple temporal scales.

Let $\mathbf{X}_{\text{clean}} \in \mathbb{R}^{C \times L}$ denote the artifact-cleaned target EEG signal (shown as $\mathbf{X}$ for simplicity) and $\hat{\mathbf{X}}$ its reconstruction. We apply random convolutional kernels organized into $G$ groups, each containing $K$ kernels. To better capture EEG temporal dynamics, DiCT employs multiple kernel sizes and dilation scales designed to model neural activity across different temporal resolutions and frequency ranges. As shown in Fig.~\ref{fig:EEG-X}, kernels within each group compete to represent local temporal patterns. The kernels operate at multiple dilation scales $d$, where $d$ controls the temporal spacing between convolutional elements. The dilation ranges and kernel configurations are selected according to EEG temporal characteristics and sampling frequencies (Details of kernel design and parameter selection are provided in Appendix~\ref{apx:DiCT}).

The convolutional responses are computed as:
\begin{equation}
\mathbf{R}^{(d)}
=
\mathbf{X} *_d \mathbf{W}^{(d)},
\quad
\mathbf{R}^{(d)} \in \mathbb{R}^{G \times K \times T},
\end{equation}
where $*_d$ denotes dilated convolution. The convolutional kernels $\mathbf{W}^{(d)}$ are shared across EEG channels and applied independently to each channel. For each group $g$ and time step $t$, the kernel with the strongest response is selected:
\begin{equation}
k^*_{g,t}
=
\arg\max_k
\mathbf{R}^{(d)}_{g,k,t}.
\end{equation}
The responses are then aggregated across time:
\begin{equation}
f^{(d)}_{g,k}
=
\sum_t
\mathbf{1}[k^*_{g,t}=k].
\end{equation}

For simplicity, we illustrate the kernel selection using hard assignments in Equations (5) and (6). However, hard argmax assignments are non-differentiable and do not allow gradients to flow through the assignment operation. Therefore, during training, we use a differentiable softmax relaxation of the assignments (detailed in Appendix~\ref{apx:DiCT_soft}).
The final representation is obtained by concatenating features across groups, dilation scales, and channels. Reconstruction loss is then computed in this structured feature space:
\begin{equation}
\mathcal{L}_{\text{DiCT}}
=
MSE(
\mathcal{F}(\mathbf{X})
-
\mathcal{F}(\hat{\mathbf{X}})).
\end{equation}

By comparing temporal patterns instead of enforcing strict point-wise alignment, DiCT improves robustness to amplitude imbalance, artifacts, and temporal misalignment while preserving multi-scale temporal structure. As shown in Fig.~\ref{fig:DiCT}, DiCT assigns a substantially lower reconstruction error to the richer mid/high-frequency reconstruction (4.26) compared to the low-frequency dominant reconstruction (11.44), indicating improved sensitivity to temporal and frequency-aware structure (additional quantitative comparisons are provided in Appendix~\ref{apx:DiCT}).

\begin{figure}
\vspace{-1.3cm}
  \centering
  \subfloat[TUH Cap 19-channel]{%
    \includegraphics[trim=1cm 0.7cm 1cm 0cm, width=0.235\linewidth]{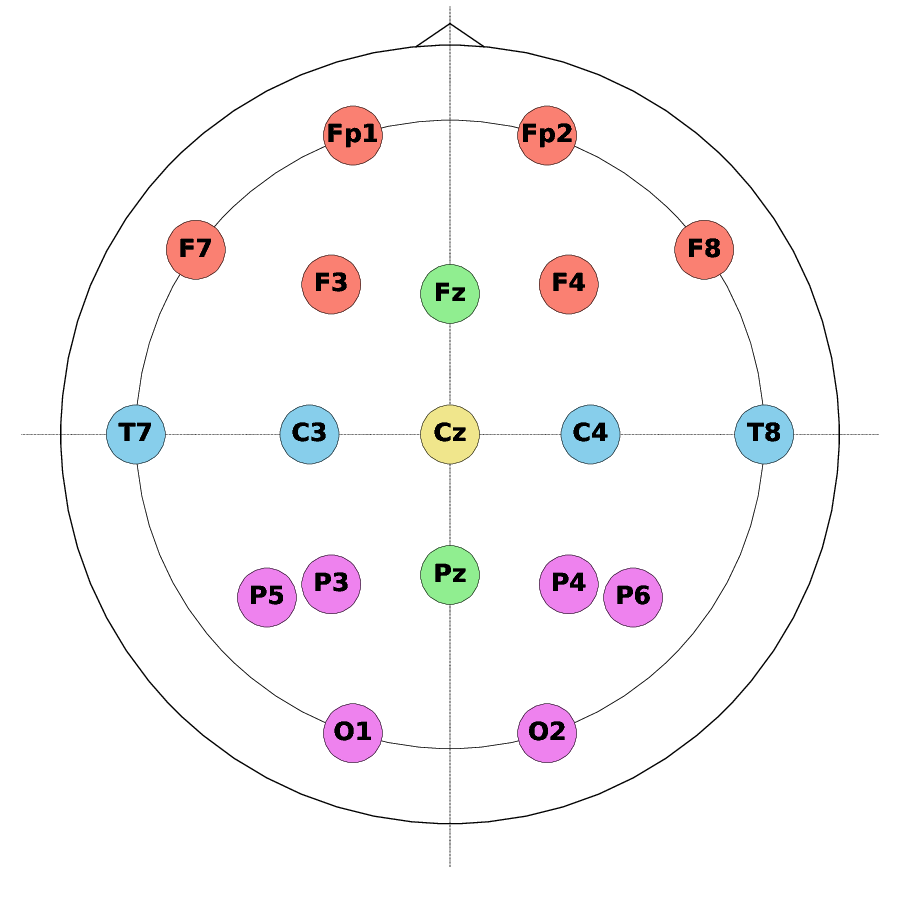}
    \label{fig:Temple_position}%
  }\hfill%
  \subfloat[Emotiv Epoc 14-channel]{%
    \includegraphics[trim=1cm 0.7cm 1cm 0cm, width=0.235\linewidth]{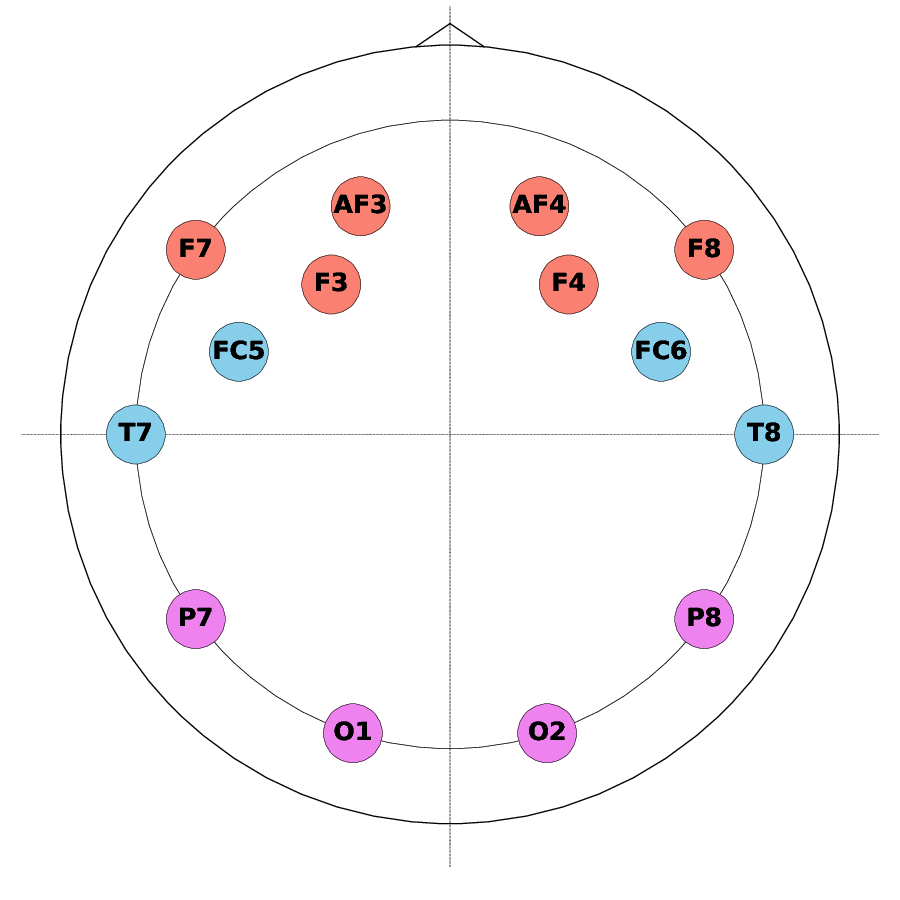}
    \label{fig:epoch_position}%
  }\hfill%
  \subfloat[BCICIV Cap 22-channel]{%
    \includegraphics[trim=1cm 0.7cm 1cm 0cm, width=0.235\linewidth]{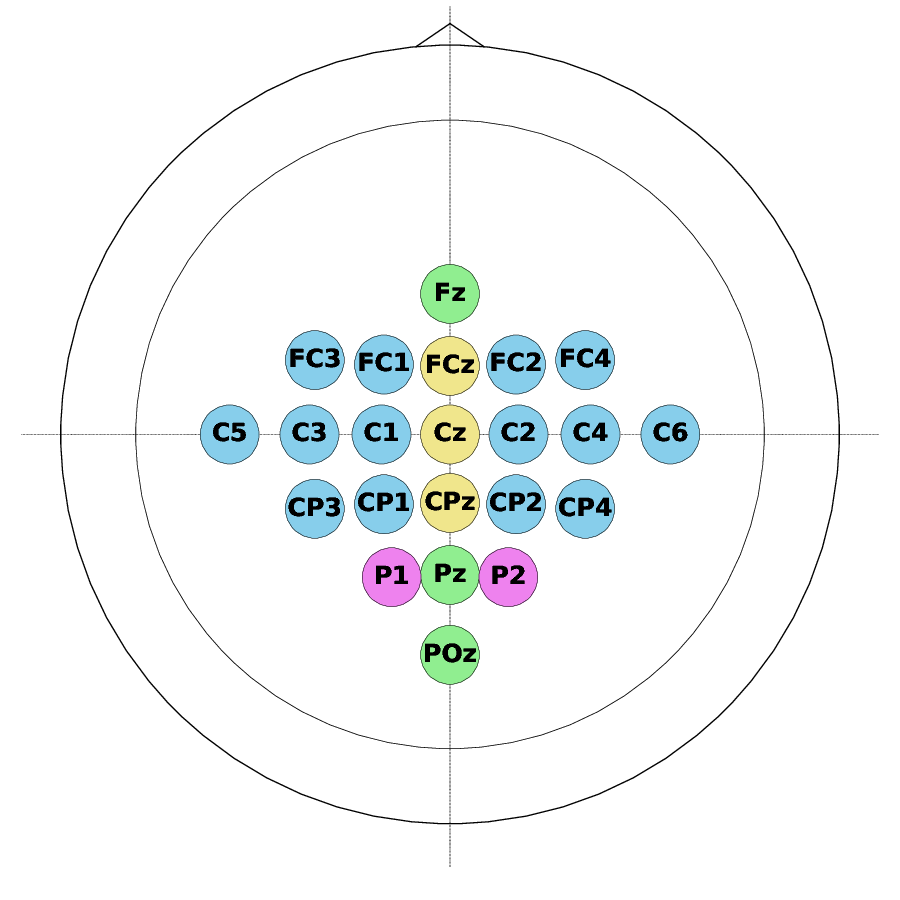}
    \label{fig:BCICIV_position}%
    }\hfill%
  \subfloat[SSVEP Cap 8-channel]{%
    \includegraphics[trim=1cm 0.7cm 1cm 0cm, width=0.235\linewidth]{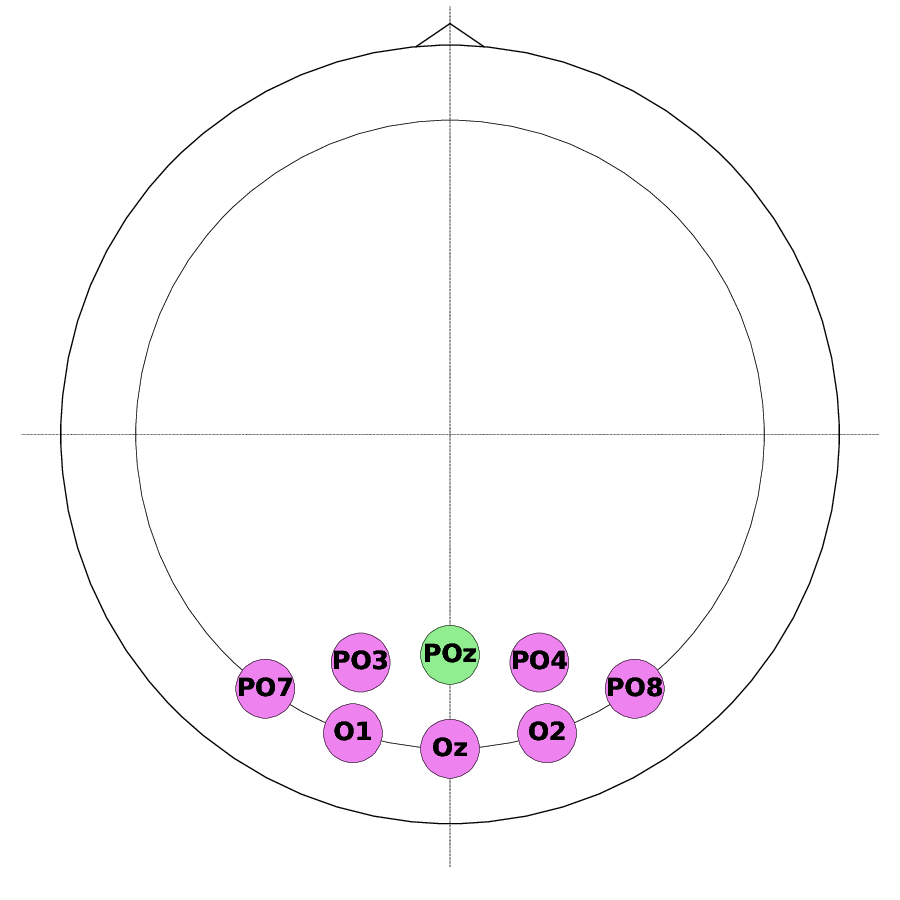}
    \label{fig:SSVEP_position}%
  }
  \caption{Comparison of four different EEG headset configurations used for data collection}
  \label{fig:headset_configs}

\end{figure}

\vspace{-0.3cm}
\section{Experimental Results}
We report the main experimental results of EEG-X, with additional setup details and supplementary analyses provided in Appendix~\ref{app:exp}.
\subsection{Datasets and Pre-processing}
We evaluate EEG-X across eight diverse datasets from five EEG headsets, ranging from portable consumer-grade devices to clinical-grade systems, in both real-world and clinical settings. We use larger datasets for in-domain evaluation and smaller ones, recorded with different headsets and channel configurations (Fig.~\ref{fig:headset_configs}), for cross-domain evaluation. A summary of the datasets is provided in Table~\ref{tbl:data}, with additional preprocessing details in Appendix~\ref{app:data}.

\begin{table}[ht]
    \centering
    \caption{Datasets for pre-training and downstream evaluation (headset configurations in Fig.~\ref{fig:headset_configs}).}
    {\footnotesize  
    \setlength\extrarowheight{1pt}
    \begin{tabular}{lcccccc}
        \hline
        \textbf{Category} & {\footnotesize \textbf{Datasets}} & \textbf{Channel} & \textbf{Length} & \textbf{\#Samples} & {\footnotesize \textbf{Task}} \\
        \hline \hline
        \multirow{4}{*}{\shortstack{\footnotesize \textbf{In-Domain}\\ \scriptsize \textbf{(Pre-training)}}} & {\footnotesize DREAMER} & 14 & 2s & 170,246 & {\footnotesize Emotion Detection} \\
        & {\footnotesize STEW} & 14 & 2s & 56,928 & {\footnotesize Mental Workload Classification} \\
        & {\footnotesize TUAB} & 19 & 10s & 409,455 & {\footnotesize Abnormal EEG Classification} \\
        & {\footnotesize TUEV\_binary} & 19 & 2s & 98,684 & {\footnotesize Epileptiform Classification}  \\
        \hline 
        \multirow{4}{*}{\shortstack{\scriptsize \textbf{Cross-Domain}\\ \scriptsize \textbf{(Downstream)}}} & {\footnotesize Crowdsourced} & 14 & 2s & 12,296 & {\footnotesize Eyes Open/Close Detection} \\
        & {\footnotesize BCICIV\_2A} & 22 & 4s & 5,184 & {\footnotesize Motor Imagery} \\
        & {\footnotesize PhysioNetMI} & 64 & 2s & 19,022 & {\footnotesize Motor Imagery} \\
        & {\footnotesize SSVEP} & 8 & 2s & 960 & {\footnotesize Steady-State Visual Evoked Potentials} \\
        
        \hline
    \end{tabular}
    } 
    \label{tbl:data}
\end{table}

\subsection{Comparison with Baseline}
We conducted comprehensive evaluations against eight state-of-the-art EEG foundation and representation learning models: MAEEG~\citep{maeeg}, BioT~\citep{Biot}, LaBraM~\citep{LaBraM}, EEGPT~\citep{EEGPT}, EEG2Rep~\citep{eeg2rep}, CBraMod~\citep{cbramod}, REVE~\citep{reve}, and ST-EEGFormer~\citep{steegformer}, using publicly available code and official implementations to ensure fair comparisons. All models use identical preprocessing pipelines, windowing strategies, and sampling rates. We report \textbf{Accuracy} (balanced), defined as the average recall across classes (\textit{Acc} for binary and \textit{B-Acc} for multi-class tasks). For binary classification, we also report \textbf{AUROC} (Area Under the ROC Curve) to summarize performance across thresholds. For multi-class tasks, we use the \textbf{Weighted F1} score, defined as the class-weighted average of F1 scores proportional to sample counts (see Appendix~\ref{app:setup} for detailed experimental settings).

\begin{table}[ht]
    \centering
    \caption{Average classification accuracy over three runs in the in-domain setting across four EEG tasks}
    {\footnotesize
    \setlength{\tabcolsep}{2.5pt}
    \begin{tabular}{l|cc|cc|cc|cc}
        \toprule
        \multirow{2}{*}{\textbf{Model}} & \multicolumn{2}{c|}{\textbf{STEW}} & \multicolumn{2}{c|}{\textbf{DREAMER}} & \multicolumn{2}{c|}{\textbf{TUAB}} & \multicolumn{2}{c}{\textbf{TUEV\_binary}} \\
        \cmidrule(lr){2-3} \cmidrule(lr){4-5} \cmidrule(lr){6-7} \cmidrule(lr){8-9}
        & \textbf{Acc} & \textbf{AUROC} & \textbf{Acc} & \textbf{AUROC} & \textbf{Acc} & \textbf{AUROC} & \textbf{Acc} & \textbf{AUROC} \\
        \midrule
        
        MAEEG   
        & 75.98 $\pm$\tiny{2.87} 
        & 81.73 $\pm$\tiny{2.95} 
        & 52.36 $\pm$\tiny{3.12} 
        & 54.45 $\pm$\tiny{2.98} 
        & 77.56 $\pm$\tiny{1.45} 
        & 86.56 $\pm$\tiny{1.67} 
        & 85.75 $\pm$\tiny{2.03} 
        & 91.36 $\pm$\tiny{1.89} \\

        BioT    
        & 75.79 $\pm$\tiny{1.12} 
        & 86.48 $\pm$\tiny{1.45} 
        & 52.85 $\pm$\tiny{1.34} 
        & 54.19 $\pm$\tiny{1.23} 
        & 79.21 $\pm$\tiny{1.56} 
        & 87.42 $\pm$\tiny{1.78} 
        & 86.28 $\pm$\tiny{1.43} 
        & 93.06 $\pm$\tiny{1.67} \\

        LaBraM  
        & 75.26 $\pm$\tiny{1.35} 
        & 84.37 $\pm$\tiny{1.67} 
        & 55.85 $\pm$\tiny{1.45} 
        & 55.34 $\pm$\tiny{1.23} 
        & 80.61 $\pm$\tiny{1.54} 
        & 88.77 $\pm$\tiny{1.76} 
        & 85.43 $\pm$\tiny{1.48} 
        & 90.10 $\pm$\tiny{1.59} \\

        EEGPT   
        & 77.65 $\pm$\tiny{2.95} 
        & 85.98 $\pm$\tiny{3.01} 
        & 54.11 $\pm$\tiny{3.12} 
        & 54.91 $\pm$\tiny{2.87} 
        & 78.91 $\pm$\tiny{1.49} 
        & 87.16 $\pm$\tiny{1.85} 
        & 84.05 $\pm$\tiny{2.03} 
        & 90.26 $\pm$\tiny{1.78} \\

        EEG2Rep 
        & 76.71 $\pm$\tiny{1.23} 
        & 85.89 $\pm$\tiny{1.45} 
        & 57.09 $\pm$\tiny{1.32} 
        & 56.35 $\pm$\tiny{1.21} 
        & 80.21 $\pm$\tiny{1.54} 
        & 88.43 $\pm$\tiny{1.67} 
        & 89.83 $\pm$\tiny{1.38} 
        & 95.57 $\pm$\tiny{1.47} \\ 

        CBraMod 
        & 75.89 $\pm$\tiny{1.43} 
        & 84.89 $\pm$\tiny{1.37} 
        & 56.09 $\pm$\tiny{1.09} 
        & 55.56 $\pm$\tiny{1.21} 
        & 80.05 $\pm$\tiny{1.40} 
        & 88.63 $\pm$\tiny{1.70} 
        & 87.35 $\pm$\tiny{1.42} 
        & 94.71 $\pm$\tiny{1.25} \\

        REVE
        & 79.95 $\pm$\tiny{1.28}
        & 86.72 $\pm$\tiny{1.41}
        & 56.84 $\pm$\tiny{1.17}
        & 56.14 $\pm$\tiny{1.24}
        & \underline{80.48 $\pm$\tiny{1.45}}
        & \underline{88.82 $\pm$\tiny{1.62}}
        & \underline{90.21 $\pm$\tiny{1.36}}
        & \underline{95.83 $\pm$\tiny{1.31}} \\

        ST-EEGFormer
        & \underline{80.02 $\pm$\tiny{1.36}}
        & \underline{88.11 $\pm$\tiny{1.42}}
        & \underline{57.21 $\pm$\tiny{1.25}}
        & \underline{56.58 $\pm$\tiny{1.20}}
        & 79.88 $\pm$\tiny{1.39}
        & 88.02 $\pm$\tiny{1.55}
        & 88.05 $\pm$\tiny{1.41}
        & 94.63 $\pm$\tiny{1.28} \\

        \midrule
        
        \rowcolor{blue!10} 
        \textbf{EEG-X}   
        & \textbf{80.32 $\pm$\tiny{1.21}} 
        & \textbf{88.37 $\pm$\tiny{1.43}} 
        & \textbf{58.14 $\pm$\tiny{1.35}} 
        & \textbf{57.64 $\pm$\tiny{1.29}} 
        & \textbf{81.02 $\pm$\tiny{1.48}} 
        & \textbf{88.97 $\pm$\tiny{1.62}} 
        & \textbf{91.15 $\pm$\tiny{1.41}} 
        & \textbf{96.18 $\pm$\tiny{1.39}}  \\

        \bottomrule
    \end{tabular}
    }
    \label{tab:indomain}
\end{table}

\textbf{In-domain evaluation:}
Table~\ref{tab:indomain} reports average classification accuracy over three runs compared with other state-of-the-art methods. \textbf{Bold} indicates the highest performance for each dataset, while \underline{underlined} values denote the second-best results, consistent across all tables in this paper. In the in-domain evaluation, each model is pre-trained and fine-tuned on the same dataset, ensuring evaluation within a consistent recording setup for each task.

Overall, EEG-X achieves the best average performance across all EEG tasks. Recent methods such as REVE and ST-EEGFormer also demonstrate strong performance across multiple datasets, frequently achieving second-best results. This highlights the effectiveness of spatially-aware channel representations and reconstruction-based pretraining for EEG foundation models. In particular, latent-space reconstruction methods such as EEG2Rep generally outperform direct raw-space reconstruction approaches like MAEEG and EEGPT on noisy datasets such as TUAB and TUEV\_binary, suggesting the importance of noise-robust representation learning. EEG-X further improves upon these methods by reconstructing artifact-cleaned EEG signals together with the proposed DiCT-enhanced reconstruction objective, enabling the model to learn noise-robust representations while preserving informative temporal and frequency-aware structure from the input signals.

\textbf{Cross-domain evaluation as a foundation model:}
Table~\ref{tab:cross-domain} presents the average classification accuracy of EEG-X, highlighting its role as a \emph{foundation model} for EEG representation learning. We compare EEG-X against recent transformer-based EEG representation learning models that employ channel-wise tokenization and are therefore capable of handling cross-domain scenarios with varying channel configurations between pre-training and downstream tasks (note that MAEEG and EEG2Rep are excluded as they rely on convolutional tokenizers). In this setup, all models are pre-trained solely on the TUAB dataset, while the pre-trained tokenizer and encoder are frozen and only a lightweight linear classifier is trained on top of the learned representations.

\begin{table}

    \centering
    \caption{Cross-domain performance comparison of models pre-trained on TUAB dataset.}
    {\footnotesize
    \setlength{\tabcolsep}{1.5pt}
        \begin{tabular}{l|cc|cc|cc|cc}
            \toprule
            \multirow{2}{*}{\textbf{Model}} 
            & \multicolumn{2}{c|}{\textbf{Crowdsourced}} 
            & \multicolumn{2}{c|}{\textbf{BCICIV\_2A}} 
            & \multicolumn{2}{c|}{\textbf{SSVEP}}
            & \multicolumn{2}{c}{\textbf{PhysioNetMI}} \\

            \cmidrule(lr){2-3} 
            \cmidrule(lr){4-5} 
            \cmidrule(lr){6-7}
            \cmidrule(lr){8-9}

            & \textbf{Acc} & \textbf{AUROC} 
            & \textbf{B-Acc} & \textbf{Weighted F1} 
            & \textbf{B-Acc} & \textbf{Weighted F1}
            & \textbf{B-Acc} & \textbf{Weighted F1} \\

            \midrule

            BioT    
            & 77.76 $\pm$\tiny{1.12} 
            & 83.62 $\pm$\tiny{1.34} 
            & 27.26 $\pm$\tiny{1.05} 
            & 25.26 $\pm$\tiny{1.23} 
            & 29.92 $\pm$\tiny{1.41} 
            & 27.27 $\pm$\tiny{1.36}
            & 48.51 $\pm$\tiny{2.70}
            & 59.03 $\pm$\tiny{2.44} \\

            LaBraM  
            & 77.28 $\pm$\tiny{1.25} 
            & 89.83 $\pm$\tiny{1.47} 
            & 32.44 $\pm$\tiny{1.13} 
            & 27.37 $\pm$\tiny{1.21} 
            & 29.69 $\pm$\tiny{1.28} 
            & 27.41 $\pm$\tiny{1.19}
            & 47.41 $\pm$\tiny{1.60}
            & 57.92 $\pm$\tiny{1.48} \\

            EEGPT   
            & 79.55 $\pm$\tiny{1.36} 
            & 87.79 $\pm$\tiny{1.52} 
            & 30.07 $\pm$\tiny{1.17} 
            & 24.68 $\pm$\tiny{1.08} 
            & 26.06 $\pm$\tiny{1.11} 
            & 27.40 $\pm$\tiny{1.07}
            & 46.96 $\pm$\tiny{2.10}
            & 57.37 $\pm$\tiny{1.95} \\

            CBraMod 
            & 78.41 $\pm$\tiny{1.31} 
            & 88.73 $\pm$\tiny{1.45} 
            & 30.07 $\pm$\tiny{1.12} 
            & 24.68 $\pm$\tiny{1.09} 
            & 28.24 $\pm$\tiny{1.34} 
            & 27.79 $\pm$\tiny{1.25}
            & 47.88 $\pm$\tiny{3.10}
            & 58.26 $\pm$\tiny{2.87} \\

            REVE
            & \underline{83.92 $\pm$\tiny{1.29}}
            & 90.72 $\pm$\tiny{1.38}
            & \underline{35.91 $\pm$\tiny{1.11}}
            & \underline{33.88 $\pm$\tiny{1.16}}
            & \underline{31.18 $\pm$\tiny{1.35}}
            & \underline{27.94 $\pm$\tiny{1.27}}
            & 50.33 $\pm$\tiny{1.43}
            & 60.88 $\pm$\tiny{1.31} \\

            ST-EEGFormer
            & 81.47 $\pm$\tiny{1.27}
            & \underline{91.42 $\pm$\tiny{1.41}}
            & 34.16 $\pm$\tiny{1.09}
            & 31.74 $\pm$\tiny{1.12}
            & 30.65 $\pm$\tiny{1.31}
            & 27.36 $\pm$\tiny{1.24}
            & \underline{50.61 $\pm$\tiny{1.35}}
            & \underline{61.26 $\pm$\tiny{1.29}} \\

            \midrule

            \rowcolor{blue!10} 
            \textbf{EEG-X} 
            & \textbf{86.89 $\pm$\tiny{1.28}} 
            & \textbf{93.51 $\pm$\tiny{1.41}} 
            & \textbf{37.58 $\pm$\tiny{1.19}} 
            & \textbf{35.32 $\pm$\tiny{1.15}} 
            & \textbf{32.97 $\pm$\tiny{1.42}} 
            & \textbf{28.15 $\pm$\tiny{1.33}}
            & \textbf{51.89 $\pm$\tiny{1.20}}
            & \textbf{62.31 $\pm$\tiny{1.08}} \\

            \bottomrule
        \end{tabular}
        } \vspace{-0.3cm}
    \label{tab:cross-domain}

\end{table}

As shown in Table~\ref{tab:cross-domain}, EEG-X consistently outperforms existing methods across all downstream tasks. Interestingly, recent methods such as REVE, which also incorporate spatially-aware channel embeddings, demonstrate stronger cross-domain performance compared to earlier approaches based purely on learnable channel embeddings. This observation further highlights the importance of incorporating electrode location information when training foundation models on heterogeneous EEG datasets. The TUAB pre-training dataset contains 19 channel configurations, while downstream tasks involve 14-, 22-, 64-, and 8-channel EEG setups with limited channel overlap. By leveraging location-based channel embeddings, EEG-X preserves spatial relationships among neighboring electrodes and enables robust generalization across different devices, channel counts, and recording configurations.

\subsection{Ablation study}
\textbf{Channel Embedding}:
We evaluate the impact of channel embeddings on model performance in both in-domain and cross-domain scenarios by comparing three configurations: without channel embedding, with learnable channel embedding, and our proposed location-based channel embedding in EEG-X. As shown in Table~\ref{tab:ablation}, incorporating channel embeddings improves the average performance compared with removing them in both settings. In the in-domain setting, learnable channel embedding improves the average score from 76.50 to 77.07, while the full EEG-X model achieves 77.66. The improvement is more pronounced in the cross-domain setting, where the average increases from 47.46 without channel embedding to 47.53 with learnable channel embedding and 52.33 with the full EEG-X model. These results highlight the importance of incorporating spatial information into channel representations. In particular, our location-based channel embedding provides additional benefits in cross-domain scenarios, potentially helping the model preserve spatial relationships among neighboring electrodes and generalize across datasets with different channel configurations.

\textbf{DiCT}: We further investigate the contribution of the Dictionary Convolution Transformation (DiCT) by comparing the full EEG-X model against a variant without DiCT. Removing DiCT reduces the average performance from 77.66 to 74.54 in the in-domain setting and from 52.33 to 49.34 in the cross-domain setting. The performance degradation is observed across all four datasets in both settings. These results demonstrate the importance of DiCT in learning effective EEG representations. By incorporating dictionary-based convolutional transformations into the reconstruction objective, DiCT helps the model capture meaningful temporal and spectral patterns, contributing to more discriminative representations and improved generalization across recording domains.

\textbf{Artifact Removal Raw Space}:
We examine the impact of artifact-aware reconstruction by comparing reconstruction of the original raw EEG signals against reconstruction of the rASR-cleaned targets used in EEG-X. Specifically, we compare the full EEG-X model, which reconstructs artifact-cleaned signals, with a variant that uses the same architecture and DiCT objective but reconstructs the original raw signals ($\mathcal{L}_{\text{rec (Raw)}}$). As shown in Table~\ref{tab:ablation}, raw-space reconstruction consistently underperforms the full EEG-X model across all four in-domain and cross-domain datasets. The average performance decreases from 77.66 to 75.24 in the in-domain setting and from 52.33 to 50.20 in the cross-domain setting. These results suggest that using artifact-cleaned EEG signals as reconstruction targets helps the model learn more robust and meaningful representations by reducing the influence of artifacts and non-neural noise during self-supervised training.

\begin{table}[t]
    \centering
    \caption{Ablation study of EEG-X components. The average column
    reports the mean score across datasets, with the difference from
    full EEG-X in parentheses.}
    \label{tab:ablation}
    {\small
    \setlength{\tabcolsep}{5pt}
    \renewcommand{\arraystretch}{1.12}
    \begin{tabular}{l|cccc|c}
        \toprule
        \multicolumn{6}{c}{\textbf{In-Domain}} \\
        \midrule
        \textbf{Component} & \textbf{STEW} & \textbf{DREAMER}
        & \textbf{TUAB} & \textbf{TUEV\_binary} & \textbf{Average} \\
        \midrule
        Without channel embedding
            & 79.44 & 56.17 & 79.94 & 90.46
            & 76.50 ($\downarrow$ 1.16) \\
        Learnable channel embedding
            &80.07 & 57.12 & 80.94 & 90.14
            & 77.07 ($\downarrow$ 0.59) \\
        Without DiCT
            & 76.91 & 54.36 & 78.88 & 88.01
            & 74.54 ($\downarrow$ 3.12) \\
        Raw reconstruction
            & 77.90 & 55.41 & 79.41 & 88.22
            & 75.24 ($\downarrow$ 2.42) \\
        \textbf{Full EEG-X}
            & \textbf{80.32} & \textbf{58.14} & \textbf{81.02}
            & \textbf{91.15} & \textbf{77.66} \\
        \midrule
        \multicolumn{6}{c}{\textbf{Cross-Domain}} \\
        \midrule
        \textbf{Component} & \textbf{Crowdsourced}
        & \textbf{BCICIV\_2A} & \textbf{SSVEP}
        & \textbf{PhysioNetMI} & \textbf{Average} \\
        \midrule
        Without channel embedding
            & 81.64 & 33.55 & 26.82 & 47.82
            & 47.46 ($\downarrow$ 4.88) \\
        Learnable channel embedding
            & 81.65 & 33.25 & 27.00 & 48.21
            & 47.53 ($\downarrow$ 4.81) \\
        Without DiCT
            & 84.12 & 34.57 & 29.35 & 49.33
            & 49.34 ($\downarrow$ 2.99) \\
        Raw reconstruction
            & 85.34 & 35.31 & 30.02 & 50.12
            & 50.20 ($\downarrow$ 2.14) \\
        \textbf{Full EEG-X}
            & \textbf{86.89} & \textbf{37.58} & \textbf{32.97}
            & \textbf{51.89} & \textbf{52.33} \\
        \bottomrule
    \end{tabular}
    }\vspace{-0.3cm}
\end{table}

\textbf{rASR Preprocessing vs. Noise-Aware Reconstruction:}
A key question is: Why not use artifact removal (rASR) directly as a preprocessing step rather than incorporating it into the self-supervised learning process? To investigate this, we compare three settings, as shown in Table~\ref{tab:rASR_preprocess}: supervised EEG-X trained for 100 epochs using raw inputs, supervised EEG-X trained for 100 epochs using rASR-preprocessed inputs, and the self-supervised EEG-X model trained using raw inputs with rASR-cleaned reconstruction targets. The results show that applying rASR as a preprocessing step does not consistently improve downstream performance: it improves results on DREAMER and TUAB, but reduces performance on STEW and TUEV. In contrast, the self-supervised setting achieves the best performance across all four datasets. These findings suggest that directly applying artifact removal as a preprocessing step may not always preserve the signal information most useful for downstream tasks. Incorporating rASR into the self-supervised reconstruction objective instead allows the model to learn from raw inputs while using artifact-cleaned signals as reconstruction targets.

\begin{table}[ht]
    \centering
    \setlength{\extrarowheight}{3pt}
    \setlength{\tabcolsep}{4pt}
    \caption{Performance of EEG-X under different data and training settings.
    Supervised models are evaluated using raw or rASR-preprocessed inputs,
    while the self-supervised model is pretrained using raw inputs with
    rASR-cleaned reconstruction targets.}
    \label{tab:rASR_preprocess}
    {\footnotesize
    \begin{tabular}{l|cccc}
        \toprule
        \textbf{Setting} & \textbf{STEW} & \textbf{DREAMER}
        & \textbf{TUAB} & \textbf{TUEV\_binary} \\
        \midrule
        Supervised (Raw input)
            & 75.12 $\pm$\tiny{2.64}
            & 55.69 $\pm$\tiny{1.34}
            & 78.12 $\pm$\tiny{0.98}
            & 88.24 $\pm$\tiny{2.09} \\
        Supervised (rASR input)
            & 73.68 $\pm$\tiny{1.74}
            & 56.41 $\pm$\tiny{1.04}
            & 78.32 $\pm$\tiny{1.11}
            & 83.36 $\pm$\tiny{1.68} \\
        Self-Supervised (Raw input with rASR target)
            & \textbf{80.32 $\pm$\tiny{1.21}}
            & \textbf{58.14 $\pm$\tiny{1.35}}
            & \textbf{81.02 $\pm$\tiny{1.48}}
            & \textbf{91.15 $\pm$\tiny{1.41}} \\
        \bottomrule
    \end{tabular}
    \vspace{-0.3cm}
    }
\end{table}

\vspace{-0.3cm}
\section{Conclusion}

In this paper, we introduced \textit{EEG-X}, a device-agnostic and noise-robust model for EEG representation learning. EEG-X addresses two major challenges in EEG analysis: variability across recording devices and the inherently low signal-to-noise ratio of EEG signals. To tackle these challenges, EEG-X combines location-based channel embeddings for robust cross-device generalization with a noise-aware reconstruction strategy that reconstructs artifact-cleaned EEG signals instead of noisy raw inputs. We further introduced the Dictionary-based Convolution Transformation (DiCT), a lightweight structured reconstruction objective that improves robustness to amplitude variation and temporal misalignment while capturing multi-scale temporal patterns. Extensive experiments across diverse EEG datasets demonstrate that EEG-X consistently outperforms state-of-the-art methods in both in-domain and cross-domain settings, establishing a strong foundation model for EEG representation learning.


\bibliographystyle{plainnat}
\bibliography{Reference}
\newpage

\appendix
\section{Related Work}
A line of work tackles the variability in EEG data caused by differences in recording devices and configurations, particularly variations in channel layouts, electrode positions, and recording lengths. BIOT~\citep{Biot} addresses this challenge by tokenizing biosignals into a ``sentence-like'' structure, segmenting EEG time series in each channel into fixed-length tokens and adding temporal and channel embeddings. Similarly, EEGFormer~\citep{EEGFormer}, LaBraM~\citep{LaBraM}, and EEGPT~\citep{EEGPT} tokenize EEG signals on a channel-wise basis and process them using transformer-based architectures. LaBraM uses learnable embeddings for both channel and temporal positions, while EEGPT incorporates rotary position encodings for temporal representation alongside learnable channel embeddings. More recently, REVE~\citep{reve} introduces a 4D Fourier-based positional encoding together with learnable channel embeddings to handle varying electrode layouts and temporal lengths through coupled spatial-temporal frequency interactions, making it the closest work to our location-based channel embedding approach. Despite these advances, most existing methods either rely on learnable channel embeddings tied to specific electrode identities or require fixed spatial encoding strategies that may not generalize efficiently across heterogeneous EEG devices. In particular, learnable channel embeddings often lack explicit spatial awareness, as they do not directly reflect electrode placement on the scalp---critical for modeling meaningful EEG patterns. Moreover, the same electrode may not exist across different devices, making such embeddings impractical for handling channel mismatches and mixed-device training.

A parallel line of work has explored self-prediction approaches, such as Masked Autoencoders~\citep{mae2022}, to adapt pre-trained models for EEG~\citep{maeeg,Bendr,EEGPT,eeg2rep,cbramod,reve,steegformer}. BENDR~\citep{Bendr}, MAEEG~\citep{maeeg}, CBraMod~\citep{cbramod}, REVE~\citep{reve}, and ST-EEGFormer~\citep{steegformer} use masking and reconstruction as pretext tasks for self-supervised learning, directly reconstructing masked segments in the raw signal space. However, raw-space reconstruction is highly sensitive to noise, often leading models to preserve artifact-related patterns together with neural activity. Alternatively, LaBraM~\citep{LaBraM} uses a neural tokenizer that transforms raw EEG signals into vocabulary-based tokens, predicting tokens instead of reconstructing raw signals. However, since these approaches still rely on raw EEG representations, they remain affected by the inherently low signal-to-noise ratio of EEG, where neural activity is often buried under external artifacts and non-brain sources. EEG2Rep~\citep{eeg2rep} improves robustness against noise by introducing latent-space reconstruction instead of raw-space reconstruction, reducing sensitivity to noise and encouraging semantically rich representations through high-level feature prediction. However, it remains limited in handling datasets with varying channel configurations and recording lengths.

\section{Dictionary Convolution Transformation (DiCT)} \label{apx:DiCT}
\subsection{Motivation}
Reconstruction losses such as Mean Squared Error (MSE) are widely used in self-supervised learning and perform well in high signal-to-noise ratio (SNR) domains such as vision, language, and speech. However, they are not well aligned with the characteristics of EEG time series.

EEG signals are inherently \textit{multi-scale}, meaning that meaningful information is distributed across different temporal resolutions or frequency bands. These bands exhibit varying amplitudes and dynamics, and their relative importance depends on the underlying neural processes and recording conditions.

EEG recordings are also highly susceptible to artifacts, which often have large amplitudes and can occur across a wide range of frequencies. For example, eye blinks and movement artifacts typically dominate the low-frequency range, while muscle activity and line noise introduce strong high-frequency components. As a result, both low- and high-frequency regions of the signal can be dominated by non-neural, high-amplitude noise.

Because MSE measures point-wise differences, it is disproportionately influenced by large-amplitude components—many of which in EEG arise from artifacts—causing the model to prioritize fitting noise over meaningful neural structure.

This leads to three key limitations:

\begin{itemize}
    \item \textbf{Amplitude bias:} MSE penalizes errors proportionally to their magnitude, causing high-amplitude components—often dominated by artifacts—to drive the loss. As a result, models are encouraged to reconstruct noise or slow dominant fluctuations while ignoring informative neural activity.
    
    \item \textbf{Lack of frequency awareness:} MSE treats all point-wise deviations equally, regardless of their frequency origin. For example, failing to reconstruct a high-frequency neural oscillation may contribute little to the loss compared to a small error in a large low-frequency artifact.
    
    \item \textbf{Lack of shape awareness:} MSE compares signals point by point, making it highly sensitive to temporal misalignment. Even small shifts or phase differences can result in large penalties, despite the signals having similar structure and frequency content.
\end{itemize}

These limitations are particularly problematic for EEG, where meaningful neural activity is distributed across frequency bands, and artifacts can dominate the signal across both low- and high-frequency ranges. To address these limitations, we introduce the Dictionary Convolution Transformation (DiCT).

\subsection{Demonstrating the Limitations of Time-Domain MSE} \label{app:dict_example}

To illustrate the limitations of direct time-domain MSE and motivate DiCT, we construct a controlled example using real EEG frequency bands.

\textbf{Setup.}
Given a signal $\mathbf{X}$, we decompose it into standard EEG frequency bands using band-pass filters:
\begin{equation}
\mathbf{X} = \sum_{b \in \mathcal{B}} \mathbf{X}^{(b)},
\end{equation}
where $\mathcal{B} = \{\text{Delta}, \text{Theta}, \text{Alpha}, \text{Beta}, \text{Gamma}\}$ and $\mathbf{X}^{(b)}$ denotes the component of $\mathbf{X}$ in band $b$.

We define the bands as:
Delta (1--4 Hz), Theta (4--8 Hz), Alpha (8--14 Hz), Beta (14--30 Hz), and Gamma (30--47 Hz).

\textbf{Reconstructed signals.}
We construct three target signals:

\textit{(1) Low-frequency dominant (Delta + Theta):}
\begin{equation}
\hat{\mathbf{X}}_{\text{low}} = \mathbf{X}^{(\text{Delta})} + \mathbf{X}^{(\text{Theta})}.
\end{equation}

\textit{(2) Mid/High-frequency dominant (Theta + Alpha + Beta + Gamma):}
\begin{equation}
\hat{\mathbf{X}}_{\text{high}} = \mathbf{X}^{(\text{Theta})} + \mathbf{X}^{(\text{Alpha})} + \mathbf{X}^{(\text{Beta})} + \mathbf{X}^{(\text{Gamma})}.
\end{equation}

\textit{(3) Phase-shifted signal:}
\begin{equation}
\hat{\mathbf{X}}_{\text{shift}}(t) = \mathbf{X}(t - \Delta t),
\end{equation}
where $\Delta t$ is a small temporal shift.

\noindent
\textit{Note:} Theta is included in both reconstructions to ensure that differences arise from the relative dominance of low- versus higher-frequency components, rather than complete band removal.

\begin{figure}[h]
    \centering
    \includegraphics[trim=0cm 0cm 0cm 0cm, width=0.95\textwidth]{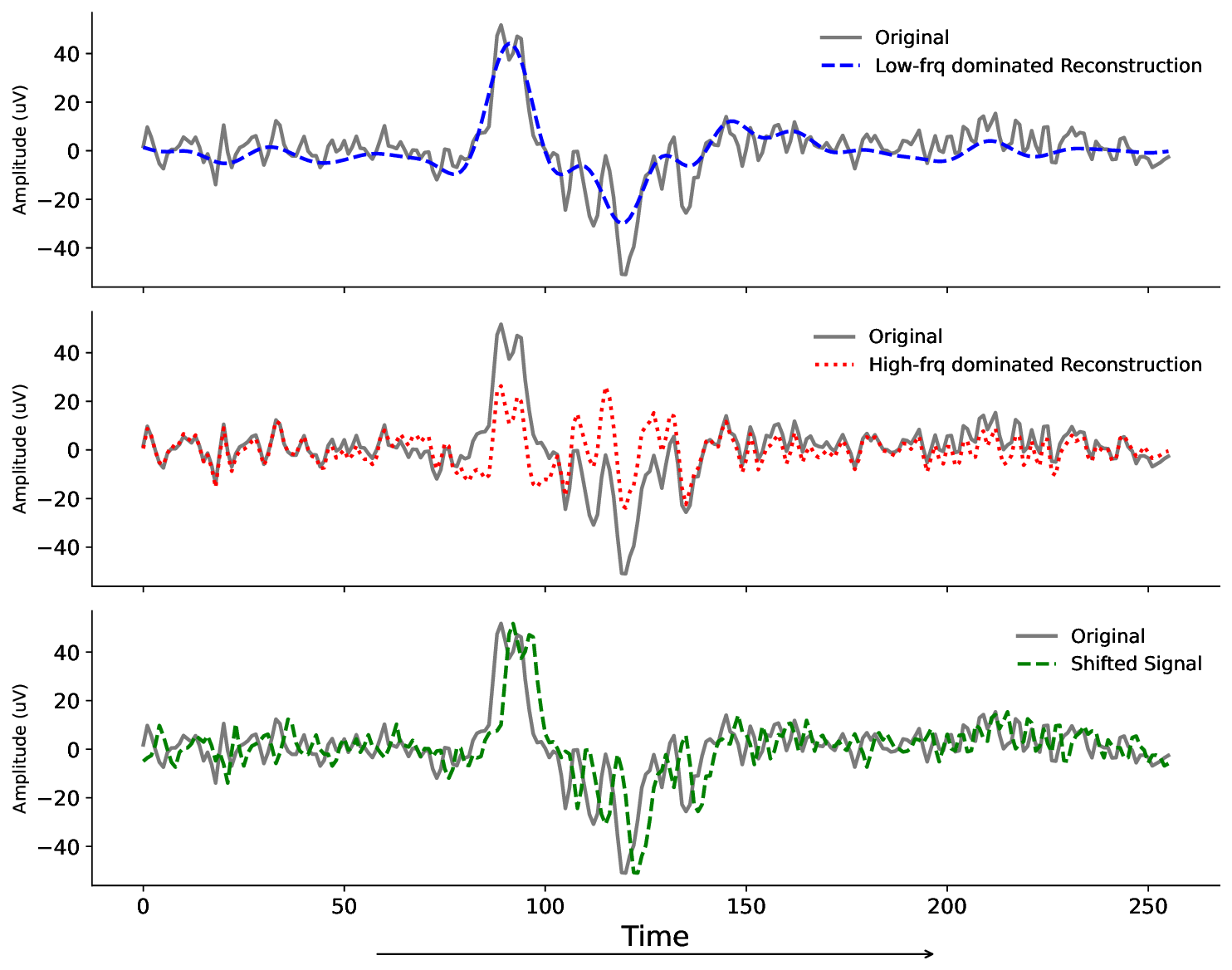}
    \caption{Original EEG signal (top) and reconstructed variants: 
    (1) low-frequency dominant (Delta + Theta), 
    (2) mid/high-frequency dominant (Theta + Alpha + Beta + Gamma), 
    and (3) phase-shifted signal. These examples highlight amplitude bias and phase sensitivity in time-domain MSE.}
    \label{fig:Apx_DiCT}
\end{figure}

Fig.~\ref{fig:Apx_DiCT} visually illustrates the original signal and the three reconstructed variants.

\textbf{Results.}
Table~\ref{tab:dict_example} reports reconstruction errors before and after applying DiCT.

\begin{table}[h]
\centering
\begin{tabular}{lcc}
\toprule
\textbf{Target} & \textbf{MSE (Time)} & \textbf{MSE (DiCT)} \\
\midrule
Low-frequency dominant & 47.55 & 11.44 \\
Mid/High-frequency dominant & 95.75 & 4.26 \\
Phase-shifted signal & 126.79 & 0.02 \\
\bottomrule
\end{tabular}
\caption{Comparison of reconstruction error before and after DiCT transformation.}
\label{tab:dict_example}
\end{table}

\textbf{Analysis.}
These results reveal a fundamental mismatch between time-domain MSE and the structural properties of EEG signals.

\textit{(i) Amplitude bias.}
Low-frequency components dominate the signal energy in EEG. Consequently, MSE is largely driven by these high-amplitude components, allowing reconstructions that discard higher-frequency structure to achieve relatively low error.

\textit{(ii) Frequency insensitivity.}
The mid/high-frequency dominant reconstruction preserves rich temporal dynamics but differs in amplitude distribution. MSE penalizes it heavily, indicating that it fails to reflect the presence of informative spectral content.

\textit{(iii) Phase sensitivity.}
A small temporal shift results in a large MSE (126.79), despite the signal being structurally identical. This highlights that MSE is sensitive to alignment rather than underlying pattern similarity.

\noindent
Importantly, MSE can assign lower error to structurally incomplete reconstructions than to those preserving richer signal content.

\textbf{Why DiCT works better.}
In contrast, DiCT significantly reduces these discrepancies. By projecting signals into a convolutional feature space, it compares the occurrence of temporal patterns rather than enforcing point-wise alignment. As a result, it (i) mitigates amplitude bias, (ii) captures multi-scale frequency content, and (iii) remains robust to temporal shifts.

Notably, the near-zero error for the shifted signal (0.02) demonstrates that DiCT preserves structural similarity under temporal misalignment—an essential property for EEG, where phase variability across trials and subjects is common.

\subsection{Parameter Selection}

The key hyperparameters in DiCT are the number of groups ($G$) and the number of kernels per group ($K$), which together define the structure of the convolutional dictionary. These parameters control the trade-off between pattern diversity and the strength of competition among patterns.

Given a fixed computational budget, increasing $G$ increases the number of independent groups, encouraging more localized and shape-preserving pattern matching. In contrast, increasing $K$ increases the number of candidate patterns within each group, improving coverage of temporal and frequency variations, but reducing the relative competition between patterns.

\textbf{Empirical evaluation.}
We evaluate several configurations of $(G, K)$ on the STEW and DREAMER datasets, as summarized in Table~\ref{tab:param_selection}.

\begin{table}[h]
\centering
\begin{tabular}{cccc}
\toprule
\textbf{Groups ($G$)} & \textbf{Kernels ($K$)} & \textbf{STEW Acc (\%)} & \textbf{DREAMER Acc (\%)} \\
\midrule
4  & 16 & 80.15 & 58.00 \\
8  & 32 & \textbf{80.32} & \textbf{58.14} \\
64 & 16 & 80.28 & 58.10 \\
8  & 64 & 80.30 & 58.12 \\
\bottomrule
\end{tabular}
\caption{Effect of the number of groups and kernels per group on downstream performance.}
\label{tab:param_selection}
\end{table}

\textbf{Analysis.}
The results highlight a clear trade-off between $G$ and $K$:

\begin{itemize}
    \item \textbf{Increasing $G$ (more groups):} promotes more structured representations by enforcing stronger competition within smaller groups, leading to improved shape sensitivity.
    
    \item \textbf{Increasing $K$ (more kernels per group):} increases pattern diversity and frequency coverage, but reduces the relative distinctiveness of patterns and increases computational cost.
\end{itemize}

We observe that extreme configurations (very large $G$ or $K$) do not yield significant improvements, indicating diminishing returns once sufficient pattern diversity is achieved. Instead, performance is maximized when balancing these two effects.

As shown in Table~\ref{tab:param_selection}, the performance is relatively stable across different configurations of (G) and (K). In particular, configurations such as $(G = 8, K = 32)$, $(G=64, K=16)$, and $(G=8, K=64)$ achieve very similar results on STEW and DREAMER. We therefore select $(G=8, K=32)$ as a practical trade-off between model configuration complexity and downstream performance, rather than as a definitive optimum.

\textbf{Kernel and dilation selection.}
To better capture EEG temporal dynamics across multiple frequency bands, DiCT employs convolutional kernels with sizes 9 and 15 together with dilation scales of $\{1,2,4,8,16\}$. These configurations are selected to model temporal patterns spanning approximately the 1--50 Hz EEG frequency range, covering Delta, Theta, Alpha, Beta, and Gamma bands under the sampling rates used in our experiments. Smaller kernels and dilations focus on high-frequency local oscillations, while larger receptive fields capture slower temporal dynamics and long-range structures.

\subsection{Differentiable Soft Assignment for DiCT}
\label{apx:DiCT_soft}

In Section~\ref{DiCT}, we introduce DiCT using hard winner assignments, where the kernel with the strongest response is selected at each time step. While this formulation provides an intuitive explanation of the dictionary-based transformation, the argmax operation is non-differentiable and prevents gradients from flowing through the assignment operation. To enable end-to-end optimization, we use a differentiable softmax relaxation during training.

Let $R_{g,k,t}^{(d)}$ denote the convolutional response for group $g$, kernel $k$, time step $t$, and dilation scale $d$. The soft assignment probability is defined as:
\begin{equation}
q_{g,k,t}^{(d)}
=
\frac{
\exp\left(R_{g,k,t}^{(d)}/\tau\right)
}{
\sum_{j=1}^{K}
\exp\left(R_{g,j,t}^{(d)}/\tau\right)
},
\end{equation}
where $\tau$ is the temperature parameter controlling the sharpness of the assignments. A lower temperature produces more concentrated assignments, whereas a higher temperature distributes probability more evenly across competing kernels. In our experiments, we use $\tau=1.0$.

The soft feature count for each kernel is then obtained by summing its assignment probabilities over time:
\begin{equation}
f_{g,k}^{(d)}
=
\sum_{t=1}^{T} q_{g,k,t}^{(d)}.
\end{equation}

Unlike hard assignments, which assign each time step exclusively to a single kernel, soft assignments distribute the contribution across competing kernels according to their relative responses. The resulting feature representation is constructed by concatenating the soft counts across EEG channels, kernel groups, kernel lengths, and dilation scales.

We apply the same fixed convolutional dictionary independently to the reconstructed EEG signal and its corresponding target signal. The dictionary kernels are shared across channels and remain fixed throughout training. Let $\mathbf{F}(\cdot)$ denote the resulting soft feature transformation. The target features are computed without gradient propagation, while gradients are retained through the reconstructed signal. The DiCT reconstruction loss is defined as:
\begin{equation}
\mathcal{L}_{\mathrm{DiCT}}
=
\operatorname{MSE}
\left(
\mathbf{F}(\hat{\mathbf{X}}),
\operatorname{stopgrad}\left(
\mathbf{F}(\mathbf{X}_{\mathrm{clean}})
\right)
\right).
\end{equation}

Here, $\hat{\mathbf{X}}$ denotes the reconstructed EEG signal and $\mathbf{X}_{\mathrm{clean}}$ denotes the corresponding processed target signal. The mean squared error is computed over the batch and all concatenated channel and dictionary features. The temporal feature counts are summed without normalization by the window length.

\subsection{Worked example of DiCT} \label{apx:DiCT-example}
\begin{figure}[!t]
  \centering
  \includegraphics[width=0.98\textwidth]{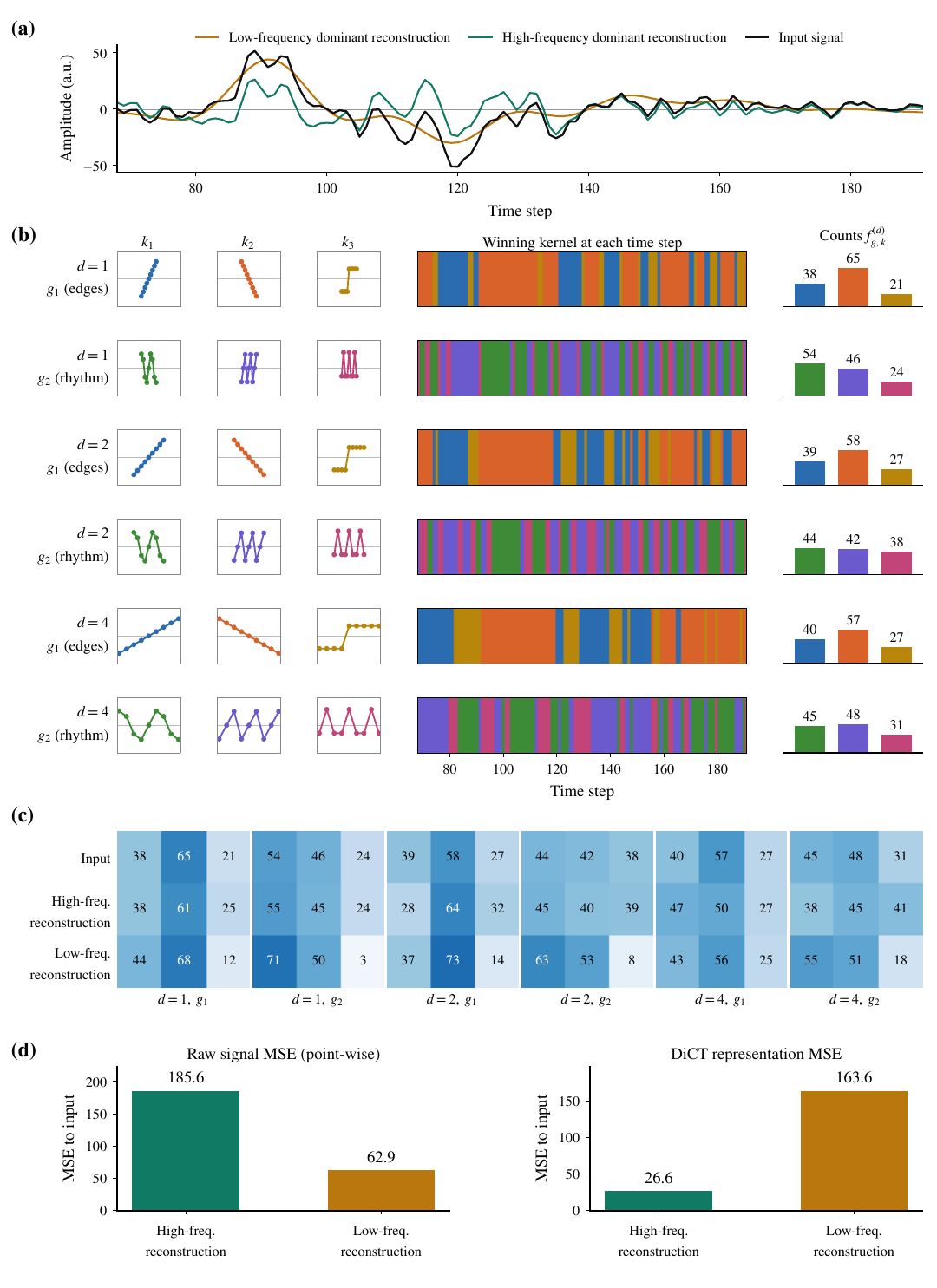}
\caption{Real example of DiCT on an EEG segment. \textbf{(a)}~Real EEG signal and two reconstructions; \textbf{(b)}~DiCT on the input EEG.
\textbf{(c)}~The representation of the input and
each reconstruction; \textbf{(d)}~The raw signal
MSE (point-wise) is lower for the low-frequency reconstruction, whereas the DiCT
representation MSE is lower for the high-frequency one, which preserves the
input's fast pattern content.}
  \label{fig:DiCT-example}
\end{figure}

Fig.~\ref{fig:DiCT-example} walks through DiCT on a 1-second EEG segment (128 time steps) and two generated targets: one dominated by high+mid frequencies and one dominated by low frequencies. Fig.~\ref{fig:DiCT-example}~(a) shows the three signals. The low-frequency target follows the large slow deflections of the input but lacks its fast fluctuations, whereas the high-frequency target reproduces the fast activity but with a distorted
large-scale shape. Fig.~\ref{fig:DiCT-example}~(b) shows DiCT applied to the input EEG signal. Two groups of three fixed kernels are used: $g_1$ contains edge-like kernels (rising ramp, falling ramp, step), and $g_2$ contains oscillatory kernels (a medium wave and a fast sine/cosine pair). Each group is applied at dilations $d\in\{1,2,4\}$, so the same kernel shape covers an increasingly long time span. At every time step the kernel with the strongest response wins, and the number of time steps won by each kernel forms the count $f^{(d)}_{g,k}$. Fig.~\ref{fig:DiCT-example}~(c) concatenates these counts into the 18-value representation $\mathcal{F}(\cdot)$ of the input and of each target; each signal is standardized to zero mean and unit variance before the transform. Fig.~\ref{fig:DiCT-example}~(d) compares the two error measures. The raw signal MSE (point-wise) prefers the low-frequency target (62.9 versus 185.6) because it matches the large-amplitude deflections that dominate the MSE. The DiCT representation MSE, the mean squared difference between the two count vectors and that of the input, prefers the high+mid frequency target (26.6 versus 163.6), because it preserves the pattern content that decides which kernels win.

For clarity, this example uses hand-picked kernels and hard winner counts;
the model uses random fixed kernels with the softmax relaxation (temperature
$\tau$). The kernels were chosen to illustrate the behavior on this segment and
are not evidence of the same ordering on all data.

\subsection{Universal Location Mapping} \label{apx:Location_mapping}
\begin{figure}[h]
    \centering
    \includegraphics[trim=1cm 1cm 1cm 1cm, width=0.8\textwidth]{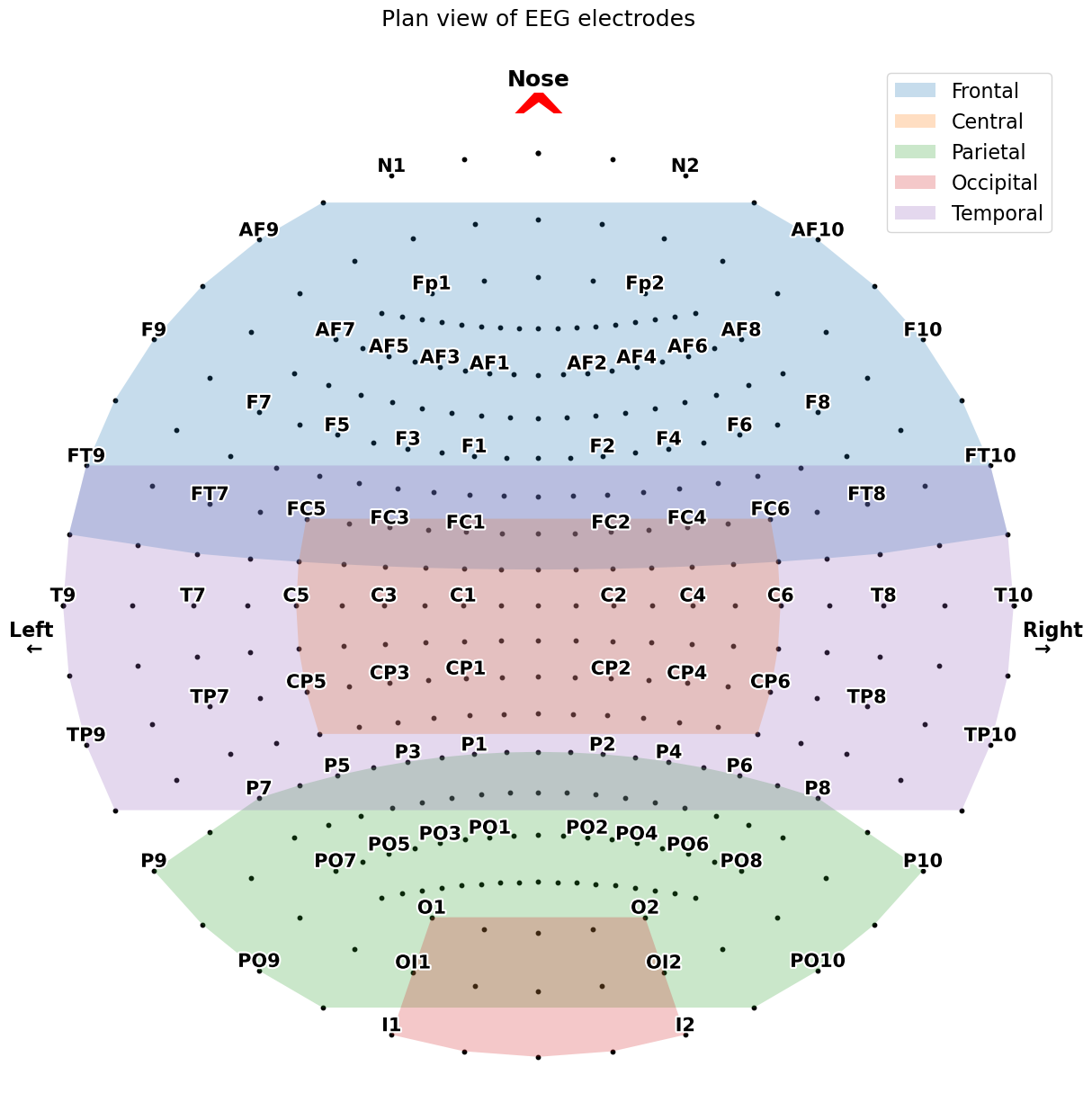}
    \caption{Dense scalp electrode mesh combining positions from the 10-05, 10-10, and 10-20 systems, used as a universal location mapping.}
    \label{fig:Location_mapping}
\end{figure}  

Fig.~\ref{fig:Location_mapping} shows a dense scalp mesh including 348 channels, covering virtually all electrode positions from the 10-05, 10-10, and 10-20 standard EEG systems. This mesh enables universal location-based channel embeddings: even if electrode names differ across devices, each sensor can be mapped to the nearest mesh point. Since typical EEG sensors are larger than 5 mm in diameter, any sensor smaller than 10 mm placed anywhere on the scalp will intersect with at least one mesh point, ensuring consistent spatial representation across devices and subjects.

\subsection{Artifact Removal Details} \label{apx:rASR}

We adopt Riemannian Artifact Subspace Reconstruction (rASR)~\citep{rASR} for artifact removal due to its robustness and suitability for large-scale EEG processing. Instead of explicitly separating sources, rASR models clean EEG activity as a low-variance subspace and removes deviations from this subspace.

\textbf{Calibration.}
rASR requires a short segment of clean EEG data. We obtain this calibration data by selecting low-variance segments using a simple signal quality criterion (e.g., RMS thresholding). Given calibration data $\mathbf{X}_{\text{cal}}$, we estimate a reference covariance matrix:
\begin{equation}
\mathbf{\Sigma}_{\text{ref}} = \frac{1}{L} \mathbf{X}_{\text{cal}} \mathbf{X}_{\text{cal}}^\top.
\end{equation}

\textbf{Artifact suppression.}
For each incoming signal window $\mathbf{X}$, we compute its covariance and compare it to the reference covariance. Deviations are measured via whitening, where directions with abnormally large variance are identified as artifacts. These directions are suppressed, and the cleaned signal is reconstructed by projecting back to the original space.

\textbf{Parameter selection.}
We tune rASR hyperparameters by injecting synthetic ERP signals into contaminated data and evaluating their recovery after denoising, ensuring a balance between artifact removal and signal preservation. We search over combinations of window length, window overlap, and rejection cutoff, and obtain the best performance with:
\[
[2,\;0.5,\;7.5] \quad \text{and} \quad [2,\;0.75,\;9],
\]
where each tuple corresponds to (window length in seconds, overlap ratio, rejection cutoff in standard deviations). Unless otherwise stated, we use the second configuration as default due to its more stable performance across datasets. For scalability, we provide a pre-trained rASR model for data cleaning, avoiding per-dataset calibration in most cases.

\textbf{Why rASR?}
Alternative artifact removal approaches include ICA-based decomposition, wavelet denoising, regression-based filtering, and spatial filtering methods such as CSP. ICA-based approaches often require explicit component decomposition and manual or semi-automatic component labeling, making them computationally expensive and less scalable for large heterogeneous EEG corpora. Wavelet and regression-based methods can effectively suppress specific artifact types but typically rely on hand-crafted assumptions or auxiliary channels such as EOG/EMG, which are not consistently available across datasets. Spatial filtering approaches are often task-specific and may not generalize well across diverse EEG paradigms. In contrast, rASR provides a lightweight and scalable artifact removal framework that does not require explicit component labeling while remaining robust across different recording conditions and EEG devices, making it particularly suitable for large-scale foundation model training.


\section{Additional Experimental Details and Results} 
\label{app:exp}

\subsection{Architecture Details} 
\label{app:arch}

EEG-X employs an asymmetric masked autoencoder architecture consisting of an 8-layer Transformer encoder and a lightweight 4-layer Transformer decoder. Similar to the standard Transformer architecture~\citep{attention2017}, we use eight attention heads in both encoder and decoder blocks to capture diverse temporal patterns from EEG signals. The embedding dimension is set to $d_e = 64$, and the feed-forward network (FFN) expands the hidden dimension by a factor of 4 before projecting it back to the original size. We follow the masking strategy of EEG2Rep~\citep{eeg2rep} (semantic subsequence-preserving masking), with a masking ratio of $75\%$. A new mask is drawn at every training step, and the encoder processes only the visible tokens.

For reconstruction learning, EEG-X uses the proposed DiCT objective, which is randomly initialized and contains no learnable parameters. EEG signals are segmented into patches of length 32 (0.25 second) with an overlap of 8 time steps, balancing temporal resolution and computational efficiency.

\subsection{Experimental Setup} 
\label{app:setup}

In our experiments, EEG-X was trained using the Adam optimizer~\citep{kingma2014adam} with a batch size of 64. The learning rate was initialized at $1 \times 10^{-3}$ and updated using a cosine learning rate scheduler~\citep{data2vec2022}. The model was pretrained for 100 epochs with early stopping based on validation loss to reduce overfitting. During in-domain fine-tuning, the tokenizer, encoder, and classification head were optimized jointly for 10 epochs using cross-entropy loss. In the cross-domain setting, the pretrained tokenizer and encoder were frozen, and lightweight linear classifier was applied on top of the learned EEG representations.

\subsection{Statistical Significance Analysis}
\label{apx:stat-test}
We performed pairwise two-sided Wilcoxon signed-rank tests on all matched results, using three runs per dataset and method, followed by Holm correction to account for multiple comparisons.

For the in-domain results in Table~\ref{tab:indomain}, EEG-X achieved the highest average rank among all evaluated methods. However, its improvements over REVE and ST-EEGFormer were not statistically significant after
Holm correction, with adjusted $p$-values of 0.5324 and 0.1700, respectively. In contrast, EEG-X significantly outperformed all other baselines, including EEG2Rep, CBraMod, LaBraM, EEGPT, BioT,
and MAEEG, with adjusted $p$-values below 0.05. The detailed adjusted $p$-values are reported in Table~\ref{tab:statistical}.

\begin{table}[t]
    \centering
    \caption{Holm-adjusted $p$-values for pairwise comparisons
    between EEG-X and competing methods.}
    \label{tab:statistical}
    {\small
    \setlength{\tabcolsep}{5pt}
    \begin{tabular}{l|cccccccc}
        \toprule
        \textbf{Setting}
        & \textbf{REVE}
        & \textbf{ST-EEGFormer}
        & \textbf{EEG2Rep}
        & \textbf{CBraMod}
        & \textbf{LaBraM}
        & \textbf{EEGPT}
        & \textbf{BioT}
        & \textbf{MAEEG} \\
        \midrule
        In-domain
        & 0.5324
        & 0.1700
        & \textbf{0.0264}
        & \textbf{0.0056}
        & \textbf{0.0036}
        & \textbf{0.0036}
        & \textbf{0.0036}
        & \textbf{0.0036} \\
        \bottomrule
    \end{tabular}
    }
\end{table}

For the cross-domain results in Table~\ref{tab:cross-domain}, EEG-X achieved the highest average rank and significantly outperformed all competing methods after Holm correction. The adjusted $p$-value was 0.0021 for all pairwise comparisons. These results demonstrate the effectiveness of EEG-X in transferring learned representations across datasets with different recording configurations and channel layouts.

\subsection{Computational Efficiency and Scalability}
\label{apx:scaling}

To study how EEG-X scales, we increase the encoder embedding width from 64 to 512 while keeping the depth (8 layers), architecture and training recipe fixed. Encoder parameters grow from 0.40M to 25.2M. We report compute as nominal \emph{dense-operation FLOP estimates}, which are hardware-independent. They count multiply-adds as 2 FLOPs and include the matrix products in the tokenizer, attention, feed-forward layers, reconstruction head and SoftDiCT convolutions. FFT, bias additions, normalization, activations, softmax, other elementwise operations, optimizer updates and data movement are excluded. Inference FLOPs cover the tokenizer, full encoder and classification head for one 2\,s EEG window (14 channels, 256 samples). Pretraining FLOPs cover one epoch (forward and backward) of the masked encoder, Predictor, decoder and reconstruction head, including both SoftDiCT branches. The pretraining batch size is 64 for all configurations. These are FLOP counts, not measured wall-clock time or hardware throughput.

\begin{table}[h]
\centering
\small
\setlength{\tabcolsep}{3pt}
\begin{tabular}{cccccccc}
\toprule
\textbf{Emb.\ dim} & \textbf{Enc.\ params} & \textbf{Total params} &
\shortstack{\textbf{Inference}\\\textbf{GFLOPs / window}} &
\shortstack{\textbf{Pretrain}\\\textbf{TFLOPs / epoch}} &
\textbf{STEW} & \textbf{DREAMER} \\
\midrule
64   & 0.40M & 0.65M & 0.114 & 44.97  & 80.32 & 58.14 \\
128  & 1.59M & 2.59M & 0.405 & 72.70  & 80.70 & 58.68 \\
256  & 6.33M & 10.28M & 1.515 & 179.87 & 80.75 & 59.12 \\
512  & 25.2M & 41.01M & 5.848 & 601.05 & 80.78 & 60.38 \\
\bottomrule
\end{tabular}
\caption{Scalability of EEG-X across encoder widths. Parameter counts and dense FLOP estimates are measured from our implementation; FLOPs are per 2\,s window (inference) or per pass over the STEW training set (39,138 windows; pretraining). Downstream columns report accuracy on STEW and DREAMER.}
\label{tab:scaling}
\end{table}

\subsection{Dataset Overview and Processing} \label{app:data}
For the Temple University Hospital (TUH) datasets, we use the default train and test splits. For the remaining datasets, we randomly select 20\% of the data subject-wise, ensuring that the same subject is not present in both the training and test sets.

\subsubsection{\textbf{Emotiv}}
We applied a bandpass filter to all Emotiv datasets and then segmented them into windows of 256 time steps, with each window corresponding to 2 seconds of recording, following EEG2Rep~\citep{eeg2rep} and KnowEEG~\citep{KnowEEG}.

\noindent\textbf{DREAMER}
DREAMER is a multimodal dataset containing electroencephalogram (EEG) and electrocardiogram (ECG) signals recorded during affect elicitation using audio-visual stimuli~\citep{dreamer}, captured with a 14-channel Emotiv EPOC headset. The dataset includes data from 23 participants, along with their self-assessments of affective states (valence, arousal, and dominance) after each stimulus. For our classification task, we focus specifically on the valence labels. The DREAMER dataset can be accessed here\footnote{\url{https://zenodo.org/records/546113}}. We use the raw EEG data, applying low-pass and high-pass filters\footnote{\url{https://torcheeg.readthedocs.io/en/v1.1.0/torcheeg.datasets.html}}. It’s important to note that for our analysis, we exclude the ECG signals and only focus on the EEG data.

\noindent\textbf{Crowdsourced}
Crowdsourced EEG data was collected while participants were engaged in a resting state task, involving periods with eyes open and eyes closed, each lasting 2 minutes. Out of 60 participants, only 13 successfully completed both conditions using 14-channel EPOC+, EPOC X, and EPOC devices. The data was initially recorded at 2048 Hz and later downsampled to 128 Hz. Raw EEG data for these 13 participants, along with preprocessing, analysis, and visualization scripts, are openly available on the Open Science Framework (OSF)\footnote{\url{https://osf.io/9bvgh/?view_only=70744f62157c46d5bd731480db1873df}}.

\noindent\textbf{Simultaneous Task EEG Workload (STEW)}
STEW dataset comprises raw EEG recordings from 48 participants using a 14-channel Emotiv EPOC headset involved in a multitasking workload experiment utilizing the SIMKAP multitasking test~\citep{stew}. Additionally, the subjects' baseline brain activity at rest was recorded before the test. The data was captured using the Emotiv Epoc device with a sampling frequency of 128Hz and 14 channels, resulting in 2.5 minutes of EEG recording for each case. Participants were instructed to assess their perceived mental workload after each stage using a rating scale ranging from 1 to 9, and these ratings are available in a separate file. Moreover, this dataset includes binary class labels, considering a workload rating of more than 4 as high and otherwise as low. We utilize these labels for our specific problem. STEW can be accessed upon request through the IEEE DataPort\footnote{\url{https://ieee-dataport.org/open-access/stew-simultaneous-task-eeg-workload-dataset}}. 

Alternatively, all the above Emotiv EEG datasets are also available on the Hugging Face platform under the \texttt{monster-monash}~\citep{monster} repository\footnote{\url{https://huggingface.co/monster-monash}} and KnowEEG~\citep{KnowEEG}.

\subsubsection{Temple University Hospital (TUH) EEG Corpus}
\noindent\textbf{TUH Abnormal EEG Corpus (TUAB)} \\
The TUH Abnormal EEG Corpus (TUAB) is a subset of the Temple University Hospital (TUH) EEG Corpus, which is one of the largest publicly available collections of clinical EEG data. The TUAB specifically focuses on EEG recordings labeled as abnormal, making it a valuable resource for studies on neurological disorders, brain function anomalies, and the development of diagnostic tools~\citep{TUAB}.

\noindent\textbf{TUH EEG Events (TUEV)} \\
The TUH EEG Events Corpus (TUEV) contains annotations of EEG segments classified into six different categories: spike and sharp wave, generalized periodic epileptiform discharges, periodic lateralized epileptiform discharges, eye movement, artifact, and background~\citep{TUEV}. For our binary classification task, we group the artifact and background categories into one class, while the remaining categories form the other class.

\noindent\textbf{Acquisition and Preprocessing}\\
The TUH Abnormal EEG Corpus (TUAB)~\citep{TUAB} and TUH EEG Events (TUEV)~\citep{TUEV} can be accessed upon request through the Temple University Electroencephalography (EEG) Resources\footnote{\url{https://isip.piconepress.com/projects/tuh_eeg/html/downloads.shtml}}. We processed both datasets to adhere to the 19-channel EEG montage following the 10-20 international system. The specific electrode configuration is as follows:

\texttt{TUH\_19\_electrodes = ['fp1', 'fp2', 'f3', 'f4', 'c3', 'c4', 'p3', 'p4', 'o1', 'o2', 'f7', 'f8', 't7', 't8', 'p6', 'p5', 'fz', 'cz', 'pz']}

\subsubsection{\textbf{BCICIV\_2A: Motor Imagery Task}:} 
The BCIC-IV-2a~\citep{BCICIV2A} dataset was recorded using a 22-channel EEG headset (Fig.\ref{fig:BCICIV_position}). The data were downsampled from 250 Hz to 128 Hz and consist of EEG recordings from 9 subjects performing four motor imagery tasks: left hand, right hand, feet, and tongue. Each subject completed two sessions on separate days, with 288 trials per session. This dataset was selected for its minimal electrode overlap with Emotiv and TUH devices (as shown in Fig.~\ref{fig:headset_configs}), providing a challenging cross-domain evaluation. 

\subsubsection{\textbf{Online SSVEP-based BCI using Riemannian Geometry}:}
The dataset comprises 8-channel EEG recordings from 12 subjects (ages 20–28) performing SSVEP-based tasks~\citep{SSVEP}. Each trial lasts 5 seconds, initiated by an auditory cue indicating which LED to focus on, or a fixation point for the reject class. The data were resampled to 128 Hz and segmented into 2-second windows. The primary task involves classifying the subject's focus on one of three LED groups or the reject class, utilizing Riemannian geometry for classification. The dataset is publicly available for research purposes.

\subsubsection{\textbf{PhysioNet Motor Imagery (PhysioNetMI)}}
The PhysioNet Motor Imagery dataset~\citep{PhysioNetMI} consists of 64-channel EEG recordings collected from 109 subjects performing different motor imagery and motor execution tasks. EEG signals were recorded using the international 10-10 electrode placement system at a sampling frequency of 160 Hz. Following previous EEG foundation model studies, we focus on the motor imagery task involving imagined left fist, right fist, both fists, and both feet movements. The data were resampled to 128 Hz and segmented into 2-second windows for training and evaluation. Due to its dense electrode configuration and large number of subjects, PhysioNetMI provides a challenging benchmark for evaluating cross-subject and cross-domain generalization in EEG representation learning.

\subsection{Visualization}
\begin{figure}[t]
  \centering
  \subfloat[epoch=5 (ACC: 65.73)]{%
    \includegraphics[width=0.30\linewidth]{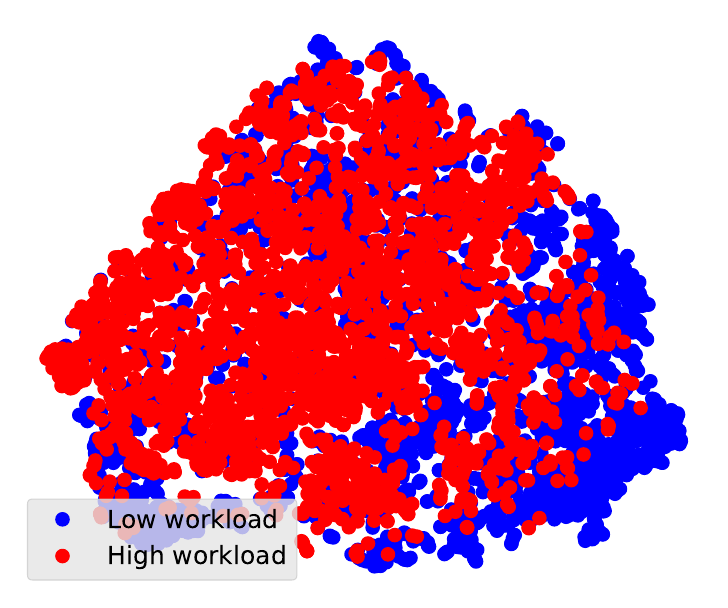}
    \label{fig:STEW_epoch_0}%
  }\hfill%
  \subfloat[epoch=10 (ACC: 67.91)]{%
    \includegraphics[width=0.30\linewidth]{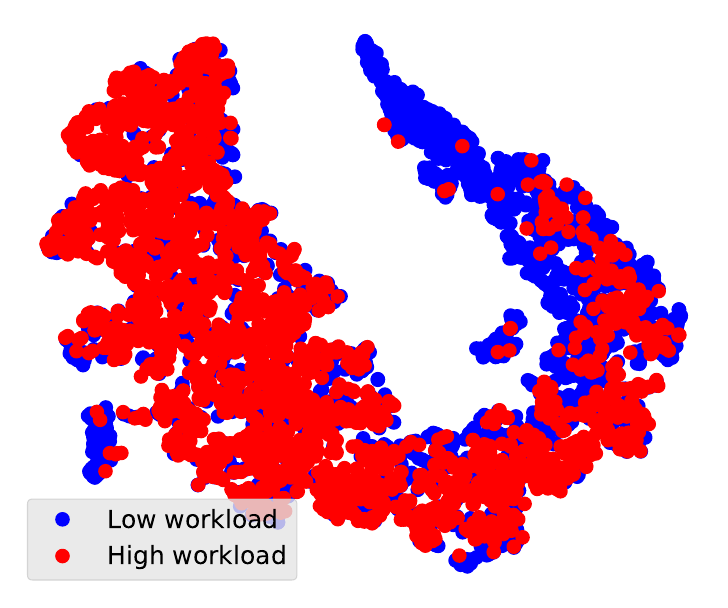}
    \label{fig:STEW_epoch_10}%
  }\hfill%
  \subfloat[epoch=25 (ACC: 71.45)]{%
    \includegraphics[width=0.30\linewidth]{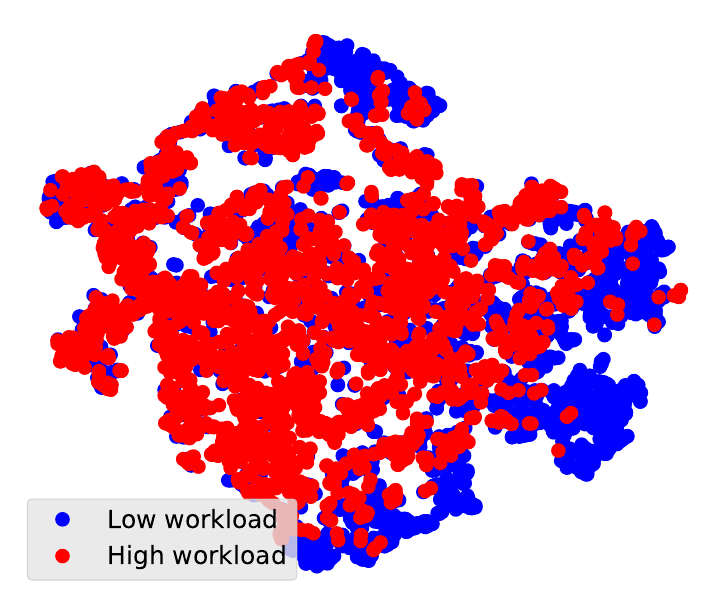}
    \label{fig:STEW_epoch_100}%
  }%

  \par\medskip

  \hspace*{\fill}%
  \subfloat[epoch=50 (ACC: 77.02)]{%
    \includegraphics[width=0.33\linewidth]{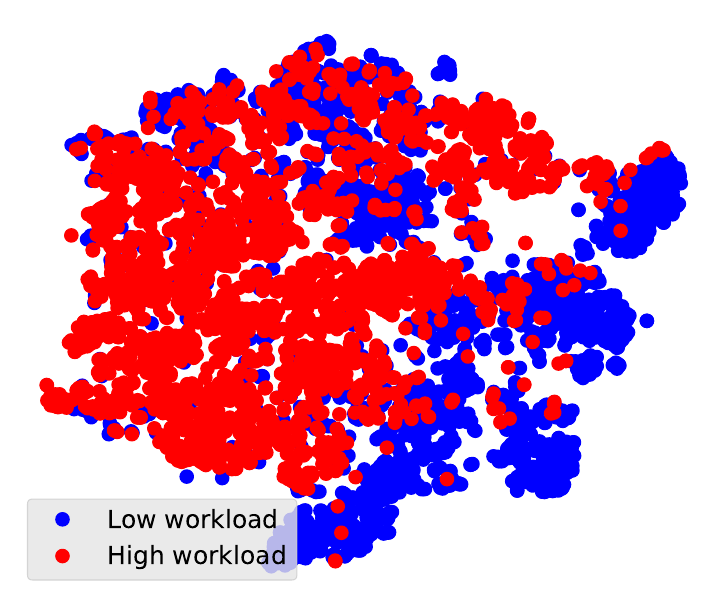}
    \label{fig:STEW_epoch_200}%
  }\hfill%
  \subfloat[epoch=100 (ACC: 78.81)]{%
    \includegraphics[width=0.33\linewidth]{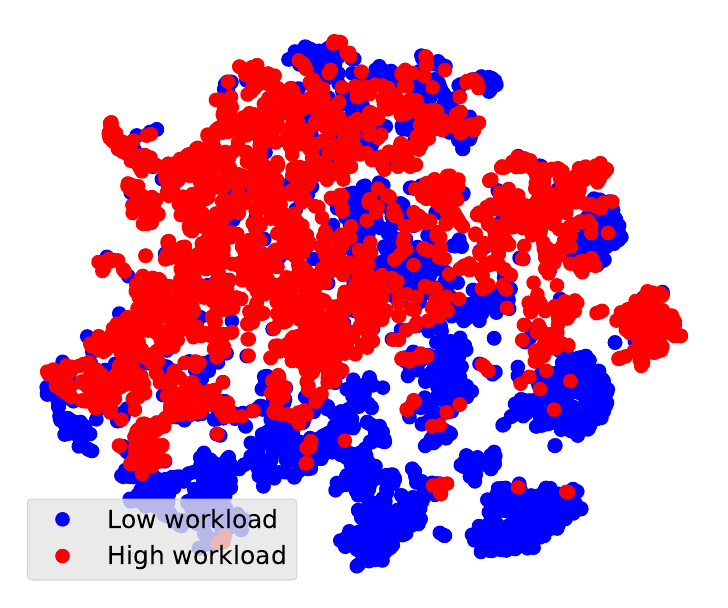}
    \label{fig:STEW_epoch_300}%
  }%
  \hspace*{\fill}%

  \caption{2D t-SNE visualization of the learned representation on the STEW dataset at different epochs during pre-training. At epoch 5, the classes are highly entangled, with separation increasing over time, becoming distinct by epoch 100. The accuracy improves from 65.73\% at epoch 5 to 78.81\% at epoch 100.}
  \label{fig:TSNE}
\end{figure}

This section visualizes the evolution of learned representations during pre-training using t-SNE plots at various epoch. As shown in Fig.~\ref{fig:TSNE} In the early stages of training (e.g., epoch 5 and 10 Fig.\ref{fig:STEW_epoch_0} and \ref{fig:STEW_epoch_10}, respectively), the embeddings from the two classes are highly entangled, with little distinction between them, showing that the model is just beginning to understand the data's structure. As training progresses, class separation improves, with noticeable distinction by epoch 25 (Fig.\ref{fig:STEW_epoch_100}) and clear separation by epoch 100 (Fig.\ref{fig:STEW_epoch_300}). Despite being unsupervised, the model leverages the STEW dataset’s inherent structure to enhance class separation. The model's accuracy also increases from 65.73\% at epoch 5 to 78.81\% at epoch 100, highlighting the benefits of self-supervised training in learning meaningful representations.

\subsection{Limitations}
\label{app:limitations}

EEG-X has several limitations. First, it is evaluated on only eight datasets, which may not fully represent the diversity of EEG devices, populations, and recording protocols. Second, its noise-aware reconstruction strategy uses rASR to generate clean EEG targets. However, rASR may not be available or practical for all datasets and recording setups, requiring alternative artifact removal methods. Third, the DiCT objective introduces additional computational overhead, which increases with the input size. Finally, our evaluation mainly focuses on classification, and its effectiveness on other EEG tasks remains to be explored.


\end{document}